\documentclass[letter, 10pt, conference]{ieeeconf}

\IEEEoverridecommandlockouts 
\usepackage{blindtext}
\usepackage{graphicx}
\usepackage{amsmath}
\usepackage{amssymb}
\usepackage{algorithm}
\usepackage[usenames,dvipsnames,svgnames,table]{xcolor}
\usepackage[noend]{algpseudocode}
\usepackage{etoolbox}
\usepackage[acronym]{glossaries} 
\usepackage[caption=false]{subfig}
\usepackage[all]{nowidow}

\makeatletter
\AfterEndEnvironment{algorithm}{\let\@algcomment\relax}
\AtEndEnvironment{algorithm}{\kern2pt\hrule\relax\vskip3pt\@algcomment}
\let\@algcomment\relax
\newcommand\algcomment[1]{\def\@algcomment{\footnotesize#1}}

\renewcommand\fs@ruled{\def\@fs@cfont{\bfseries}\let\@fs@capt\floatc@ruled
  \def\@fs@pre{\hrule height.8pt depth0pt \kern2pt}%
  \def\@fs@post{}%
  \def\@fs@mid{\kern2pt\hrule\kern2pt}%
  \let\@fs@iftopcapt\iftrue}
\makeatother

\newcommand\secref[1]{\mbox{(Sec. \ref{sec:#1})}}
\newcommand\figref[1]{\mbox{(Fig. \ref{fig:#1})}}

\newcommand\algmref[1]{\mbox{(Alg. \ref{alg:#1})}}
\newcommand\citem[1]{\mbox{\cite{#1}}}
\newcommand\seclabel[1]{\label{sec:#1}}
\newcommand\figlabel[1]{\label{fig:#1}}

\newcommand\algmlabel[1]{\label{alg:#1}}
\newcommand\core{C}
\newcommand\frontier{F}
\newcommand\free{O}
\newcommand\norm{\mathbf{e}_\mathrm{n}}

\newcommand\front{\mathbf{e}_\mathrm{f}}
\newcommand\bound{\mathbf{e}_\mathrm{b}}
\newcommand\viewp{\mathbf{x}}
\newcommand\viewo{\boldsymbol{\phi}}
\newcommand\view{\mathbf{v}:=\{\viewp,\viewo\}}
\newcommand\views{\{\viewp,\viewo\}}

\newcommand\viewn{\mathbf{v}_{i+1}}
\newcommand\frontc{\mathbf{f}}
\newcommand\frontd{\boldsymbol{s}}
\newcommand\frontdi{s}
\newcommand\cmass{\boldsymbol{\omega}}

\DeclareMathOperator*{\argmax}{arg\,max}
\DeclareMathOperator*{\argmin}{arg\,min}

\usepackage[
backend=biber,
sorting=none,
maxbibnames=99,
doi=false,
isbn=false,
url=false,
eprint=false,
giveninits=true,
style=numeric-comp,sortcites
]{biblatex}
\title{Surface Edge Explorer (SEE): \\Planning Next Best Views Directly from 3D Observations}

\author{Rowan Border$\,^{1}$, Jonathan D. Gammell$\,^{1}$ and Paul Newman$\,^{1}$
\thanks{$^{1}\,$Rowan Border, Jonathan D. Gammell and Paul Newman are with the Oxford Robotics Institute, Department of Engineering Science, Oxford University, Oxford, United Kingdom. {\tt\small rborder,gammell,pnewman@robots.ox.ac.uk}}
}

\begin{document}

\maketitle

\IEEEpeerreviewmaketitle

\newacronym{infg}{IG}{Information Gain}
\newacronym{nbv}{NBV}{Next Best View}
\newacronym{see}{SEE}{Surface Edge Explorer}

\begin{abstract}
 Surveying 3D scenes is a common task in robotics. Systems can do so autonomously by iteratively obtaining measurements. This process of planning observations to improve the model of a scene is called Next Best View (NBV) planning.

NBV planning approaches often use either volumetric (e.g., voxel grids) or surface (e.g., triangulated meshes) representations. Volumetric approaches generalise well between scenes as they do not depend on surface geometry but do not scale to high-resolution models of large scenes. Surface representations can obtain high-resolution models at any scale but often require tuning of unintuitive parameters or multiple survey stages.

This paper presents a scene-model-free NBV planning approach with a density representation. The Surface Edge Explorer (SEE) uses the density of current measurements to detect and explore observed surface boundaries. This approach is shown experimentally to provide better surface coverage in lower computation time than the evaluated state-of-the-art volumetric approaches while moving equivalent distances.
\end{abstract}

\section{Introduction}

Obtaining high-resolution 3D models of real-world scenes is a common task. These observations may be captured with a variety of robotic platforms (e.g., wheeled, articulated, aerial platforms, etc.) in a variety of different environments (e.g., outdoors, inside pipes, etc.)

The individual observations can then be combined into a single 3D representation (e.g., a triangulated 3D mesh). The quality of this model depends on how well the observations capture the scene, i.e., the number and distribution of the individual measurements. The problem of selecting and planning sensor views to obtain high-resolution models is known as \gls{nbv} planning.

\gls{nbv} planning approaches can be classified as either scene-model-based or scene-model-free. Model-based approaches \citem{Bircher2015, Kaba2016} use \textit{a priori} knowledge of the scene structure to compute a set of views from which the scene (i.e., an object or environment) is observed. These approaches work for a given scene but do not generalise well to other scenes. 

Model-free approaches often use a volumetric \citem{Connolly1985} or surface representation \citem{Hollinger2012}. Volumetric representations discretise the scene into voxels and can obtain high observation coverage with a small voxel size but do not produce high-resolution models of large scenes. Surface representations estimate surface geometry from observations and can obtain high quality models of large scenes but often require tuning of unintuitive parameters or multiple survey stages.

This paper presents the \gls{see}, a scene-model-free approach to NBV planning that uses a density representation. This representation uses a given resolution and measurement density to define a \textit{frontier} between fully and partially observed surfaces. Sensor views are proposed to observe this frontier and expand the fully observed surfaces. \gls{nbv}s are selected and new measurements are obtained until the entire scene is observed at the chosen resolution and measurement density.  

This density representation does not require an \textit{a priori} discretisation of the scene as used by volumetric approaches and scales with the number of measurements obtained and not the size of the scene. This makes \gls{see} appropriate for large-scale observations (e.g., inspecting a bridge with an aerial vehicle). \gls{see} uses a more intuitive parameterisation than many surface representations and does not require multiple survey stages.

SEE is evaluated in simulation on four standard models \citem{Krishnamurthy1996a, Turk1994, Curless1996, Newell1975} and a full-scale model of the Radcliffe Camera in Oxford \citem{Boronczyk2016} \figref{marquee}. The results show that it achieves higher surface coverage in less computational time than the evaluated state-of-the-art volumetric approaches \citem{Vasquez-Gomez2015, Kriegel2015, Delmerico2017} while requiring the sensor to travel equivalent distances.

Section II presents an overview of \gls{nbv} planning literature. Section III presents \gls{see}. Section IV presents an experimental comparison of \gls{see} with state-of-the-art volumetric approaches on four standard models and a full-scale model of the Radcliffe Camera. Sections V and VI present a discussion of the results and our plans for future work.  

\begin{figure}[tpb]
 \centering
 \subfloat[SEE]{\includegraphics[width=0.475\linewidth]{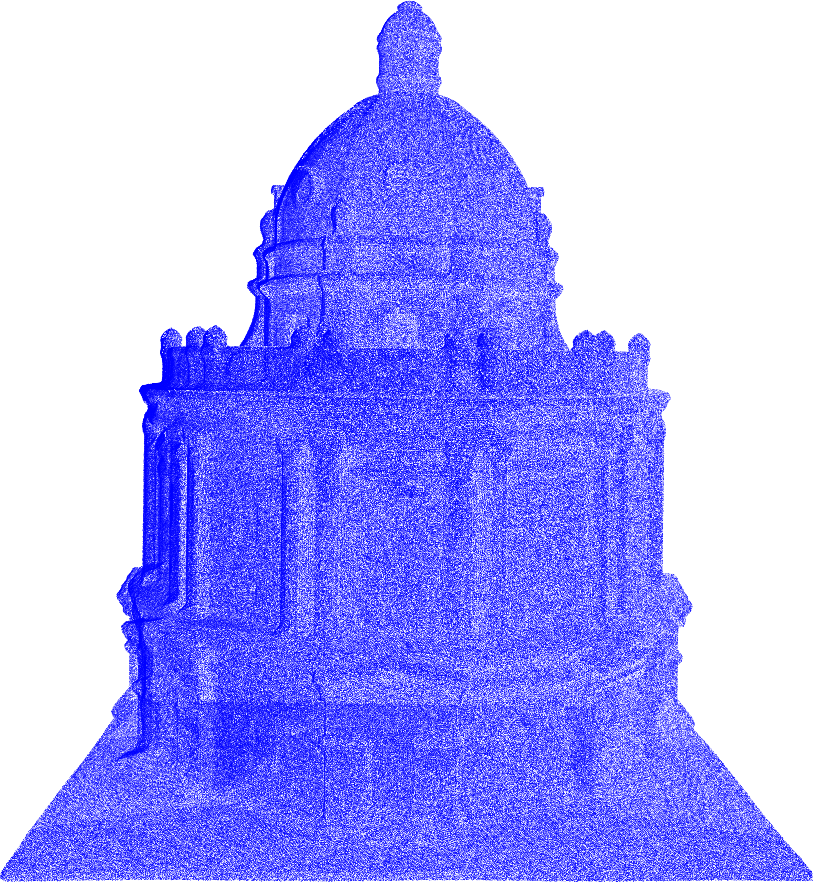}}\quad
\subfloat[AE \citem{Kriegel2015}]{\includegraphics[width=0.475\linewidth]{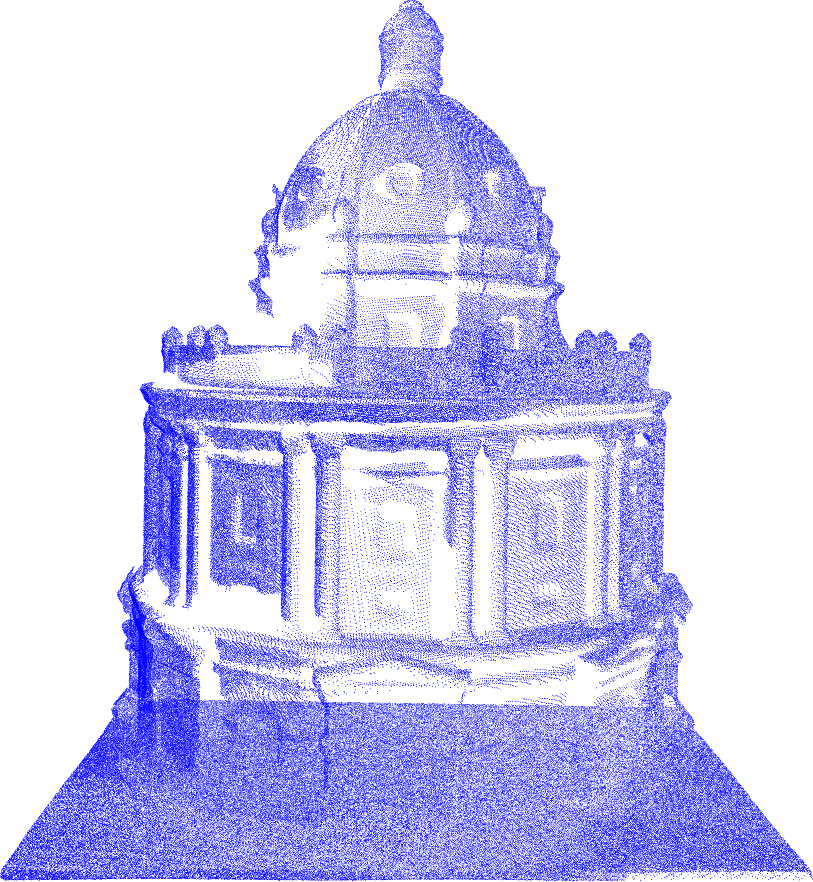}}
  \caption{A comparison of the point cloud resulting from running SEE (a) and AE \citem{Kriegel2015} (b) on a full-scale model of the Radcliffe Camera in Oxford. \gls{see} observed 99\% of the model at a $0.05$~m resolution. AE, the best-performing volumetric approach, observed 79\% in the same number of views.}
  \figlabel{marquee}
\end{figure}

\section{Related Work}

Existing NBV planning work covers a variety of scene sizes, from small objects (e.g., the Stanford Bunny \citem{Turk1994}) \cite{Vasquez-Gomez2015, Kriegel2015,  Delmerico2017, Dierenbach2016, Karaszewski2016, Khalfaoui2013, Pito1996, Yuan1995, Connolly1985} to buildings \cite{Yoder2016, MENG2017, Bircher2016, Song2017, Hollinger2012, Bissmarck2015, Bircher2015, Kaba2016, Roberts2017}.

Surveys of \gls{nbv} planning literature \citem{Tarabanis1995, Scott2003a, Karaszewski2016a} categorise approaches based on their scene representation. The most widely used categorisation \citem{Scott2003a} classifies approaches as either scene-model-based or scene-model-free. Model-based approaches \citem{Bircher2015, Kaba2016} require an \textit{a priori} scene model and do not generalise well. Within the class of model-free approaches there are global, volumetric and surface representations.

Global representations \citem{Pito1996, Yuan1995} consider all observations as part of a single connected surface. Pito \citem{Pito1996} generates a tessellated view space and selects \gls{nbv}s to observe the boundaries of a partial mesh until the mesh boundaries are closed. It obtains high-resolution models but requires a fixed work-space and known sensor model. Yuan \citem{Yuan1995} estimates the geometry of surface patches and selects views to observe the unknown space between them and obtain a single surface but only demonstrates it on simple surface geometries.

Volumetric representations \citem{Vasquez-Gomez2015, Delmerico2017, Connolly1985, Yoder2016, Bircher2016, Song2017, MENG2017, Bissmarck2015} discretise a bounded scene volume into a voxel grid from which view selection metrics can be computed. Seminal work by Connolly \citem{Connolly1985} uses a metric that counts the number of unseen voxels visible from potential views on a tessellated sphere encompassing the scene. View metrics in later work \citem{Vasquez-Gomez2015, Delmerico2017} consider multiple factors but still sample views from a tessellated surface. Vasquez-Gomez et al. \citem{Vasquez-Gomez2015} rank potential views based on reachability, distance, overlap with previous views and the number of visible unseen voxels. Delmerico et al. \citem{Delmerico2017} use \gls{infg} metrics to evaluate views based on voxel visibility, observability and proximity to existing observations. 

The model resolution obtained from a volumetric representation depends on the resolution of the voxel grid and the number of potential views. Smaller voxels and more potential views allow for greater model detail but require higher computational costs to raytrace each view. These representations are difficult to scale to large scenes without lowering the model quality or increasing the computation time.

Volumetric representations \citem{Yoder2016, Bircher2016, Song2017, MENG2017, Bissmarck2015} have been applied to large scenes despite these limitations. Most approaches mitigate the increase in computation time by reducing the number of potential views. Yoder et al. \citem{Yoder2016} only sample views to observe the frontier between seen and unseen voxels and select \gls{nbv}s with a view selection metric that balances view utility and travel distance. Meng et al. \cite{MENG2017} similarly only sample views that observe frontier voxels and select \gls{nbv}s with an \gls{infg} metric. Bircher et al. \citem{Bircher2016} use the RRT algorithm \citem{LaValle1998} to plan paths through known voxels and sample views at the vertices of the RRT tree to observe unknown voxels. The \gls{nbv} is selected from the sampled views with an \gls{infg} metric. Song et al. \citem{Song2017} present a similar approach to \citem{Bircher2016} using the RRT* algorithm \citem{Karaman2011} to plan a path to the \gls{nbv} that maximises the observation of frontier voxels. Potential views are sampled within a given radius of the RRT* path and the subset that provides the greatest coverage is selected.  

Reducing the number of potential views can mitigate the increased computational cost of large scenes but the resolution of the voxel grid is still limited by the raytracing complexity. Bissmarck et al. \citem{Bissmarck2015} compare raytracing algorithms that consider voxel observability, frontier voxels, sparse ray casting and using a hierarchy of voxel grid resolutions to reduce this complexity. They demonstrate that these algorithms outperform simple raycasting in terms of computation time but a \gls{nbv} planning approach using the algorithms for view selection is not presented.

Surface representations \citem{Dierenbach2016, Khalfaoui2013, Roberts2017, Hollinger2012} estimate surface geometry from sensor observations (e.g., by triangulating measurements into a mesh) and compute views to extend the surface boundaries and improve the surface quality. Some approaches incrementally extend the surface representation with new observations \citem{Dierenbach2016, Khalfaoui2013} while others use a multistage survey to iteratively refine a surface model of the scene \citem{Roberts2017, Hollinger2012}. 

Dierenbach et al. \citem{Dierenbach2016} estimate surface geometry by training a neural network to generate a simplified mesh from sensor measurements. Point density is computed locally around the mesh vertices and views are proposed to extend the mesh and obtain a given point density. Khalfaoui et al. \citem{Khalfaoui2013} apply density-based clustering to sensor observations and propose views to observe the cluster boundaries until the maximum distance between cluster centers is below a given threshold. These approaches can obtain high-resolution models but require tuning of unintuitive parameters.

Multistage approaches \citem{Roberts2017, Hollinger2012} refine an existing surface mesh that is often obtained manually or with a preplanned path. Hollinger et al. \citem{Hollinger2012} represent the mesh uncertainty as a Gaussian process and propose views to improve the surface estimation. Roberts et al. \citem{Roberts2017} sample potential views within a given distance of the mesh surface, select the minimal  subset that can provide complete coverage and plan the shortest path between them.

Some work \citem{Kriegel2015, Karaszewski2016} presents approaches using both volumetric and surface representations. Kriegel et al. \citem{Kriegel2015} combine a volumetric representation with an \gls{infg} view selection metric and a surface representation that selects views to extend the boundaries of a surface mesh and obtain a given point density. Karaszewski et al. \citem{Karaszewski2016} obtain an initial scene survey with a volumetric representation and then fill discontinuities in the observed surfaces based on the local point density. The local measurement density is also considered by \gls{see} but without the complexity of using a different underlying representation. 

\vspace{1ex}

\gls{see} is a \gls{nbv} planning approach that uses a density representation. Unlike volumetric representations, it scales well to large scenes and is shown to obtain accurate and complete models of scenes at any scale (i.e., both \emph{bunnies} and \emph{buildings}). Unlike surface representations, it does not require multistage surveys or have unintuitive parameters. SEE instead uses only measurement density and resolution.

\section{Surface Edge Explorer (SEE)}

SEE seeks to observe an entire scene with a minimum measurement density. This measurement density is defined by the resolution, $r$ and target density, $\rho$, used to detect frontiers in the measurements. Frontiers are detected by classifying sensor measurements (i.e., points) based on the number of neighbouring points within the distance $r$. Points with sufficient neighbours (i.e., the local density is greater than or equal to $\rho$) are classified as \emph{core} and those without are classified as \emph{outliers}. Outlier points with both core and outlier neighbours are then classified as \emph{frontier} points \figref{class}. These frontier points represent the boundary between fully and partially observed surfaces \secref{front_det}.

The scene observation is expanded by taking measurements at these frontiers. Potential views are proposed by estimating the local surface geometry around frontier points as a plane described by a set of orthogonal vectors \figref{surf}. These vectors describe the normal to the local surface, the density boundary and the direction of partial observation (i.e., the frontier) \secref{surf_geom}.

Views are proposed orthogonal to this locally estimated surface plane to maximise sensor coverage \figref{vp}. The view distance can be specified by the user or defined as a function of the sensor parameters and desired resolution \secref{view_prop}.

The \gls{nbv} is selected from these \emph{view proposals} to reduce the distance from both the current sensor position and the first observation of the scene. This guides observations to expand one frontier at a time and decreases the total distance travelled by the sensor \secref{nbv}.

The proposed views will not expand frontiers on discontinuous or highly non-planar surfaces. These views are iteratively adjusted in response to new observations until the frontier point is observed or a sufficient number of attempts have been made to classify it as an outlier. Points classified as outliers will not be reprocessed unless a new point is observed nearby \secref{view_adj}.

SEE continues to select NBVs until there are no more frontier points and all measurements have been classified as core or outlier points. This can be achieved in unbounded real-world problems by discarding all measurements outside of a predefined scene boundary \secref{term}.

\begin{figure}[tpb]
 \centering
  \includegraphics[width=\linewidth]{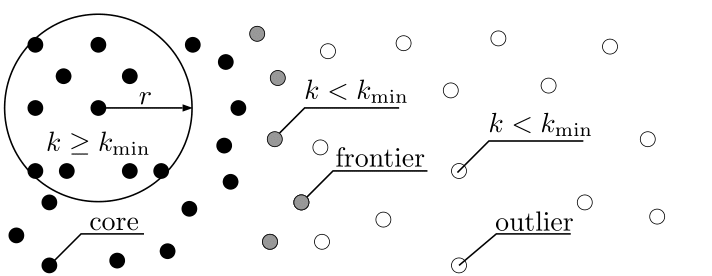}
  \caption{An illustration of SEE's density-based classification. Points with a sufficient number of neighbours are classified as core points (black) while those without are outlier points (white). Points with both core points and outlier points in their neighbourhood are frontier points (grey).}
  \figlabel{class}
\end{figure}

\subsection{Frontier Detection}
\seclabel{front_det}

\begin{figure}[tpb]
 \centering
  \includegraphics[width=\linewidth]{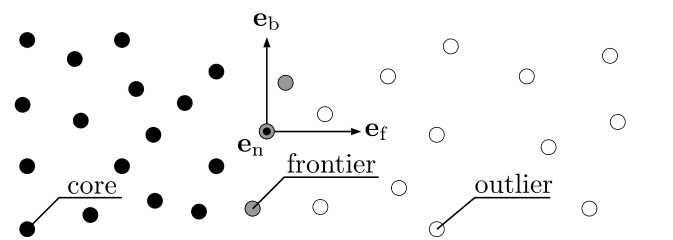}
  \caption{An illustration of SEE's local surface geometry estimation. The geometry of the surface at the frontier points (grey) is estimated from nearby points with an orthogonal set of vectors. These vectors are orientated normal to the surface, $\norm$ (out of the page), parallel to the boundary line, $\bound$ and perpendicular to the boundary line (i.e., into the frontier), $\front$.}
  \figlabel{surf}
\end{figure}

Frontiers between fully and partially observed surfaces are detected by performing density-based classification of sensor measurements (i.e., points). Points are classified as either core, frontier or outlier based on the number of neighbouring points, $k$, with a radius, $r$, of the point \figref{class}. The number of observed points in the $r$-ball is compared with the minimum number of points, $k_\mathrm{min}$, necessary to satisfy the desired point density, $\rho$, where $k_\mathrm{min} = \frac{4}{3}\rho\pi r^3$.

This density-based classification approach is based on DBSCAN \citem{Ester1996}. DBSCAN classifies a set of sensor measurements, $P := \{\mathbf{p}_i\}_{i=1}^n$ where $\mathbf{p}_i \in \mathbb{R}^3$, as core points, $\core$, frontier points, $\frontier$, or outlier points, $\free$. These labels are complete and unique such that \[P \equiv \core \cup \frontier \cup \free \quad\mathrm{and}\quad \core \cap \frontier \equiv \core \cap \free \equiv \frontier \cap \free \equiv \emptyset\,.\]

A point is classified as a core point if it has more than $k_\mathrm{min}$ neighbours within a distance $r$, \[\core := \{\mathbf{p} \in P \;|\; |N_\mathbf{p}| \geq k_\mathrm{min}\}\,,\]
where $N_\mathbf{p}$ is the set of points within $r$ of $\mathbf{p}$, \[N_\mathbf{p} := N(P,r,\mathbf{p}) := \{\mathbf{q} \in P \;|\;||\mathbf{q} - \mathbf{p}|| \leq r\}\,,\]
$||\cdot||$ is the $\mathrm{L}^2$-norm and $|\cdot|$ is set cardinality.

A point is classified as a frontier point if it is not a core point but has both core and outlier neighbours, 
\[\frontier := \{\mathbf{p} \in P \;|\; |N_\mathbf{p}| < k_\mathrm{min} \;\land\; N_\mathbf{p} \;\cap\; \core \not= \emptyset \;\land\; N_\mathbf{p} \;\cap\; \free \not= \emptyset\}\,.\]
It is otherwise classified as an outlier point,
\[\free = P \setminus (\core \cup \frontier)\,.\]

This paper modifies DBSCAN to classify measurements obtained from incremental observations \algmref{inc_dbscan_alg}. When a new sensor observation is obtained, the set of new measurements, $M$, is combined with the existing classification sets, $\core$, $\frontier$ and $\free$ (Line 1). Each new point, $\mathbf{p}$, is processed and added to either the core, frontier or outlier point sets (Line 3). Any new point that has not yet been classified is added to the (re)classification queue, $Q$, along with its neighbourhood points (Lines 4--5). If a point in the queue is not a core point then it is (re)classified based on the new measurements (Lines 6--7). Points with insufficient neighbours to be core are classified as frontier points if they have both core and outlier neighbours or otherwise as outlier points (Lines 9--14). Points with sufficient neighbours are classified as core points (Line 16). If the point was previously unclassified then its neighbourhood is added to the (re)classification queue and it is marked as classified (Lines 19--21).

\begin{algorithm}[tpb]
\caption{POINT-CLASSIFIER($M, \core, \frontier, \free, r, k_\mathrm{min}$)}
\algmlabel{inc_dbscan_alg}
\begin{algorithmic}[1]
\small
\State {$P := \core \cup \frontier \cup \free \cup M$}
\State {$V \gets \emptyset$}
\ForAll{{$\mathbf{p} \in M$}}
\If{{$\mathbf{p} \notin V$}}
\State $Q \gets N(P,r,\mathbf{p}) \cup \{\mathbf{p}\}$
\ForAll{$\mathbf{q} \in Q$}
\If{{$\mathbf{q} \notin C$}}
\State $N_\mathbf{q} \gets N(P,r,\mathbf{q})$
\If{$|N_\mathbf{q}| < k_\mathrm{min}$}
\If{{$N_\mathbf{q} \cap C \neq \emptyset$} \textbf{and} {$N_\mathbf{q} \cap O \neq \emptyset$}}
\State $\frontier \gets \frontier \cup \{\mathbf{q}\}$
\State $\free \gets \free \setminus \{\mathbf{q}\}$
\Else
\State $\free \gets \free \cup \{\mathbf{q}\}$
\EndIf
\Else
\State $\core\gets \core\cup \{\mathbf{q}\}$
\State $\frontier\gets \frontier\setminus \{\mathbf{q}\}$
\State $\free \gets \free \setminus \{\mathbf{q}\}$
\If{{$\mathbf{q} \in M$} \textbf{and} {$\mathbf{q} \notin V$}}
\State $Q \gets Q \cup N_\mathbf{q}$
\State {$V \gets V \cup \{\mathbf{q}\}$}
\EndIf
\EndIf
\EndIf
\EndFor
\EndIf
\EndFor
\end{algorithmic}
\end{algorithm}

\subsection{Surface Geometry Estimation}
\seclabel{surf_geom}

Good observations require knowledge of the surface geometry. The surface around a frontier point, $\mathbf{f}$, is approximated as locally planar through eigendecomposition of a matrix representation of its neighbourhood,
\[\mathbf{D} := [\mathbf{p}_{1}-\mathbf{f},...,\mathbf{p}_{n}-\mathbf{f}] \in \mathbb{R}^{3 \times |N_\mathbf{f}|}\,,\] where $\mathbf{p}_{i} \in N_\mathbf{f}$ are the neighbouring points.

The eigendecomposition of the square matrix, $\mathbf{A} := \mathbf{DD}^\mathrm{T}$, produces a set of eigenvalues, $\Lambda = \{\lambda_{1},\lambda_{2},\lambda_{3}\}$ and their corresponding eigenvectors, $\Upsilon = \{\mathbf{\psi}_{1}, \mathbf{\psi}_{2}, \mathbf{\psi}_{3}\}$, satisfying the eigenequation,
\[\mathbf{A}\mathbf{\psi}_i = \lambda_i\mathbf{\psi}_i\,, \;i = \{1,2,3\}\,.\]

As $\mathbf{A}$ is a real orthogonal matrix, the set of eigenvectors form an orthonormal basis (i.e., three mutually orthogonal unit vectors) of $\mathbf{D}$. Each eigenvector describes one component of the observed surface geometry \figref{surf}. The normal vector, $\norm$, is orthogonal to the surface plane. The boundary vector, $\bound$, points along the boundary between partially and fully observed surfaces. The frontier vector, $\front$, lies in the surface plane and points in the direction of partial observation.

The surface geometry components are determined sequentially from the eigenvectors, eigenvalues, view orientation and the mean of the nearby points, $\mathbf{{\bar{p}}}$, \[\mathbf{{\bar{p}}} = \frac{1}{|N_\mathbf{f}|}\sum_{\mathbf{p} \in N_\mathbf{f}}{\mathbf{p} - \mathbf{f}} \,.\] 

\subsubsection{Normal vector}
\seclabel{surf_norm_def}

The normal vector, $\norm$, is assigned as the eigenvector corresponding to the minimum eigenvalue (i.e., the direction of least surface variance), 
\[\norm = \{\mathbf{\psi}_i \;|\; \lambda_i = \min\left\lbrace\Lambda\right\rbrace\}\,.\]

The direction of the normal vector is chosen to be opposite the direction of the view orientation, $\viewo$, such that, \[|\norm \cdot \viewo| < 0\,.\]
\subsubsection{Frontier vector}
\seclabel{edge_orth_def}

The frontier vector, $\front$, is the eigenvector perpendicular to the boundary of the partially observed surface. It is assigned as the remaining eigenvector which maximises the magnitude of the dot product with the mean point,
\[\front = \argmax_\mathbf{\mathbf{\psi} \,\in\,  \Upsilon \setminus \norm} (|\mathbf{\bar{p}} \cdot \mathbf{\psi}|)\,.\]

The direction of the frontier vector is chosen to point away from the mean of the frontier point neighbourhood, into the partially observed region of the point cloud such that, \[|\front \cdot \mathbf{{\bar{p}}}| < 0\,.\]

\subsubsection{Boundary vector}
\seclabel{edge_para_def}

The remaining eigenvector is locally tangential to the boundary between the density regions and is referred to as the boundary vector. The direction of the boundary vector is given by the cross product of the normal and frontier vectors,
\[\bound := \norm \times \front\,.\]

\begin{figure}[tpb]
 \centering
  \includegraphics[width=\linewidth]{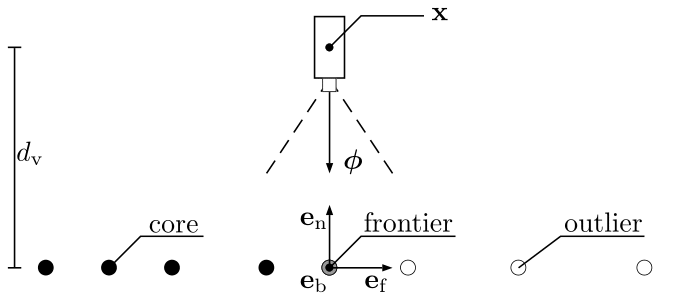}
  \caption{An illustration of SEE's initial view proposal generation. Initial view proposals, ($\viewp$, $\viewo$), are generated around each frontier point (grey) from the estimated local surface geometry, $\norm$, $
  \front$ and $\bound$. The view orientation, $\viewo$, is given by the inverse sign of the normal vector, $\viewo = -\norm$. The view position, $\viewp$, is set at a view distance, $d_\mathrm{v}$, from the frontier point in the direction of the normal vector, $\norm$. The dashed lines represent the field-of-view of the sensor. These views are adjusted when observing surfaces with discontinuities and occlusions to obtain the best view possible.}
  \figlabel{vp}
\end{figure}

\subsection{View Generation}
\seclabel{view_prop}

View proposals are generated to maximise sensor coverage of the estimated planar surface around each frontier point. A view proposal, $\view$, is defined by a view position, $\viewp$ and orientation, $\viewo$.

The view position is a distance, $d_\mathrm{v}$, on the normal vector, $\norm$, from the frontier point,
\[\viewp = \mathbf{f} + d_\mathrm{v}\norm\,.\]
The view distance may be user specified or defined as function of the sensor parameters and desired resolution.

The view orientation, $\phi$, is given by the inverse of the normal vector (i.e., pointing in the direction of the surface),
\[\viewo = -\norm\,.\]

\subsection{NBV Selection}
\seclabel{nbv}

The NBV is selected from the set of view proposals, \[W := \{\mathrm{\mathbf{g}}(\mathbf{f} \in \frontier)\}\,,\]
where $\mathrm{\mathbf{g}}$ maps frontier points to view proposals (i.e., Sec. \ref{sec:view_prop}).

SEE observes the scene while reducing total travel distance by selecting NBVs based on their \emph{incremental} and \emph{origin} distances. The incremental distance of a NBV is given by the difference between the current view position, $\viewp_i$ and the position of the proposed view. The origin distance of a NBV is given by the difference between the position of the proposed view and the first scene observation, $\viewp_0$.

The NBV, $\viewn$, is selected to minimise the global distance,
\[\viewn = \argmin_{\views \in W'}(||\viewp - \viewp_0||)\,,\]
from the set of view proposals, $W'$, within $r$ of the current view,
\[W' = \{\views \in W\;|\; ||\viewp - \viewp_i|| < r\}\,.\] 

If there are no nearby view proposals (i.e., $W' \equiv \emptyset$) then the NBV that minimises the local distance is selected,
\[\viewn = \argmin_{\views \in W}(||\viewp - \viewp_i||)\,.\]

\subsection{Local View Adjustment}
\seclabel{view_adj}

Real surfaces have discontinuities and occlusions that invalidate the locally planar assumptions and prevent expansion of the frontier. In these situations, SEE incrementally adapts the current view until either the frontier point is observed or sufficient attempts have been made to classify it as an outlier.

The locally planar assumption is often violated by surface discontinuities (e.g., edges or corners) or occlusions by other surfaces. When the frontier point is near a discontinuity, the view must be adjusted to observe both sides of it (i.e., to see around the corner). When the frontier point is occluded by another surface, the view must be adjusted to avoid the occlusion (i.e., to see around the other surface). These views are not orthogonal to the locally estimated surface. SEE attains such views by iteratively using new measurements to translate and rotate the current view to move the center of the observed points towards the frontier point.

The magnitude of the translation and rotation for each axis is determined by the displacement, $\frontd := [\frontdi_1, \frontdi_2, \frontdi_3]^T$, between the center of observed points, $\cmass$, and the frontier point along the axis,
\[\frontd = \mathbf{R}^\mathrm{T}_\mathrm{d}(\frontc - \cmass)\,,\]
where $\mathbf{R}_\mathrm{d} = [\norm\;\front\;\bound]$ is a rotation into a local frame.

The view is first translated along the frontier vector by a distance, $d_\mathrm{f}$,
\[d_\mathrm{f} = \frontdi_1(d_\mathrm{t} + 1)\,,\]
and rotated around the boundary vector by $\theta_\mathrm{b}$,
\[\theta_\mathrm{b} = \tan^{-1}\left(\frac{d_\mathrm{v}\frontdi_1d_\mathrm{t}}{d^2_\mathrm{v} + \frontdi_1^2(d_\mathrm{t} + 1)}\right)\,.\]

It is then translated along the boundary vector by a distance, $d_\mathrm{b}$,
\[d_\mathrm{b} = \frontdi_2(d_\mathrm{t} + 1)\,,\]
and rotated around the frontier vector by $\theta_\mathrm{f}$,
\[\theta_\mathrm{f} = \tan^{-1}\left(\frac{d_\mathrm{v}\frontdi_2d_\mathrm{t}}{d^2_\mathrm{v} + \frontdi_2^2(d_\mathrm{t} + 1)}\right)\,.\]

The distance factor, $d_\mathrm{t}$, determines the magnitude of the translation and rotation for the view adjustment. SEE scales it exponentially with the number of view adjustments, $n$, for a given frontier point, $d_\mathrm{t} = 2^n$. This stops the size of the view adjustment from converging to zero as the center of observed points moves closer to the frontier point.

The position and orientation of the adjusted view, $\viewn$, is then given by,
\begin{align*}
\viewp_{i+1} &= \frontc - d_\mathrm{v}\viewo_{i+1}\,,\\
\viewo_{i+1} &= \frac{\frontc - \mathbf{R}_\mathrm{f}\mathbf{R}_\mathrm{b}(\viewp_i + d_\mathrm{f}\front + d_\mathrm{b}\bound)}{||\mathbf{R}_\mathrm{f}\mathbf{R}_\mathrm{b}(\viewp_i + d_\mathrm{f}\front + d_\mathrm{b}\bound) ||}\,.
\end{align*}

The rotation matrices, $\mathbf{R}_\mathrm{b}$ and $\mathbf{R}_\mathrm{f}$, are computed with Rodrigues' rotation formula \citem{RodriguesO.1840a} using the frontier and boundary axes and angles, $\theta_\mathrm{f}$ and $\theta_\mathrm{b}$,
\begin{align*}
\mathbf{R}_\mathrm{b} &= (\cos \theta_\mathrm{b})\mathbf{I} + \sin \theta_\mathrm{b}\bound^\wedge + (1 - \cos \theta_\mathrm{b}) \bound\bound^\mathrm{T} \,,\\
\mathbf{R}_\mathrm{f} &= (\cos \theta_\mathrm{f})\mathbf{I} + \sin \theta_\mathrm{f}\front^\wedge + (1 - \cos \theta_\mathrm{f}) \front\front^\mathrm{T} \,,
\end{align*}
where,
\[\mathbf{u}^{\wedge} = \begin{bmatrix}
    0       & -u_{2} & u_{1} \\
    u_{2}       & 0 & -u_{0} \\
    -u_{1}       & u_{0} & 0
\end{bmatrix}\,,\]
and $\mathbf{I}$ is the identity matrix.

The sensor is moved to the adjusted view and another observation is obtained. This process is repeated iteratively until the frontier is expanded (i.e., the other side of the surface discontinuity is observed) or the Euclidean distance between the frontier point and the center of observed points stops reducing. If this termination criterion is reached then the view is reinitialised on the viewing axis from which the frontier point was observed (i.e., where no occluding surface exists) but at a distance from the surface no greater than that of the observing view, $\viewp_\mathrm{obs}$.

This new view position is 
\[\viewp_{i+1} = \frontc - \min\{||\frontc - \viewp_\mathrm{obs}||\,,\, d_\mathrm{v}\}\, \viewo_{i+1}\,.\]

The new view orientation is 
\[\viewo_{i+1} = \frac{\frontc - \viewp_\mathrm{obs}}{||\frontc - \viewp_\mathrm{obs}||}\,.\]

When starting the view adjustment from the observation viewing axis, the distance factor is reinitialised, $d_\mathrm{t} = 1$, and adjustment is again performed until termination. If this process also reaches the termination criterion then the frontier point is reclassified as an outlier point.

\subsection{Completion}
\seclabel{term}

\gls{see} completes the observation of a scene when the final frontier point has been observed and all points are classified as either core points or outliers. This termination criterion assumes that the observable scene is finite. In the real world this condition can be met by defining a scene boundary and discarding all measurements outside it.

\section{Evaluation}

\begin{figure*}[tpb]
\centering
\captionsetup[subfigure]{}
\captionsetup[subfigure]{labelformat=empty}
\captionsetup[subfigure]{}
\subfloat[Stanford Armadillo ($1$~m) \citem{Krishnamurthy1996a}]{\includegraphics[width=.24\linewidth]{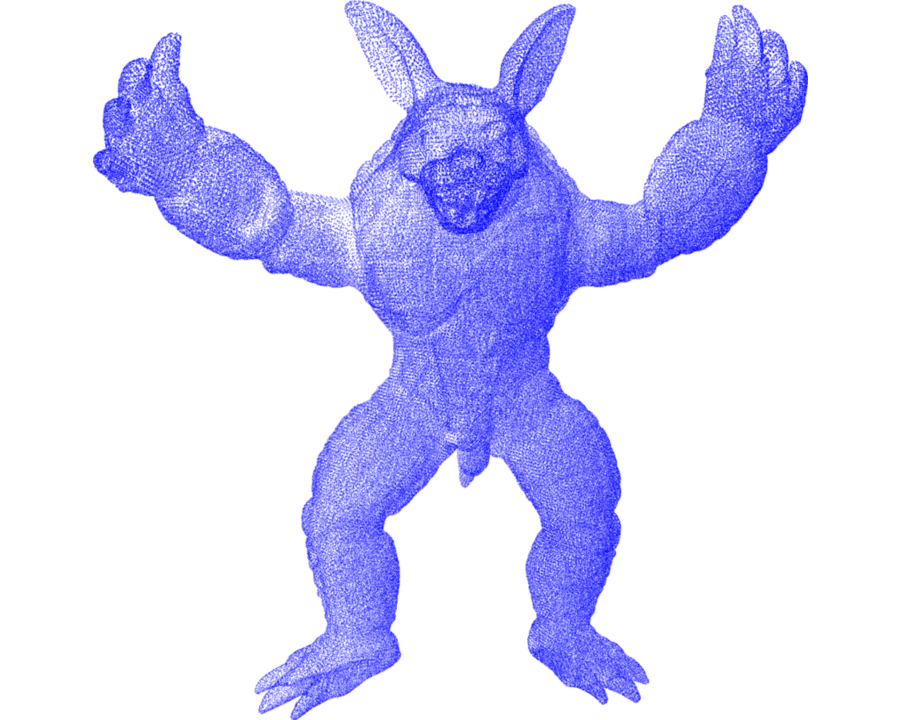}} \hfill
\captionsetup[subfigure]{labelformat=empty}
\subfloat[]{\includegraphics[width=.24\linewidth]{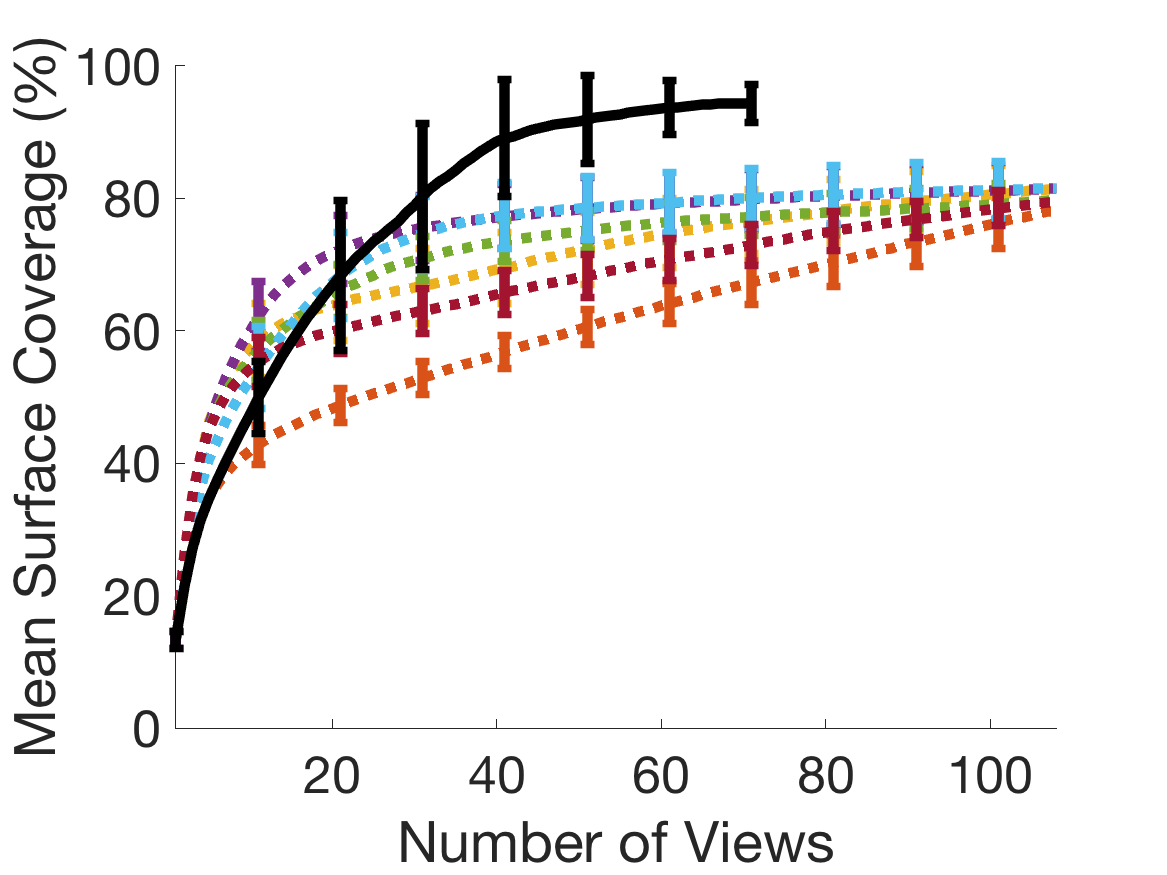}} \hfill
\subfloat[]{\includegraphics[width=.24\linewidth]{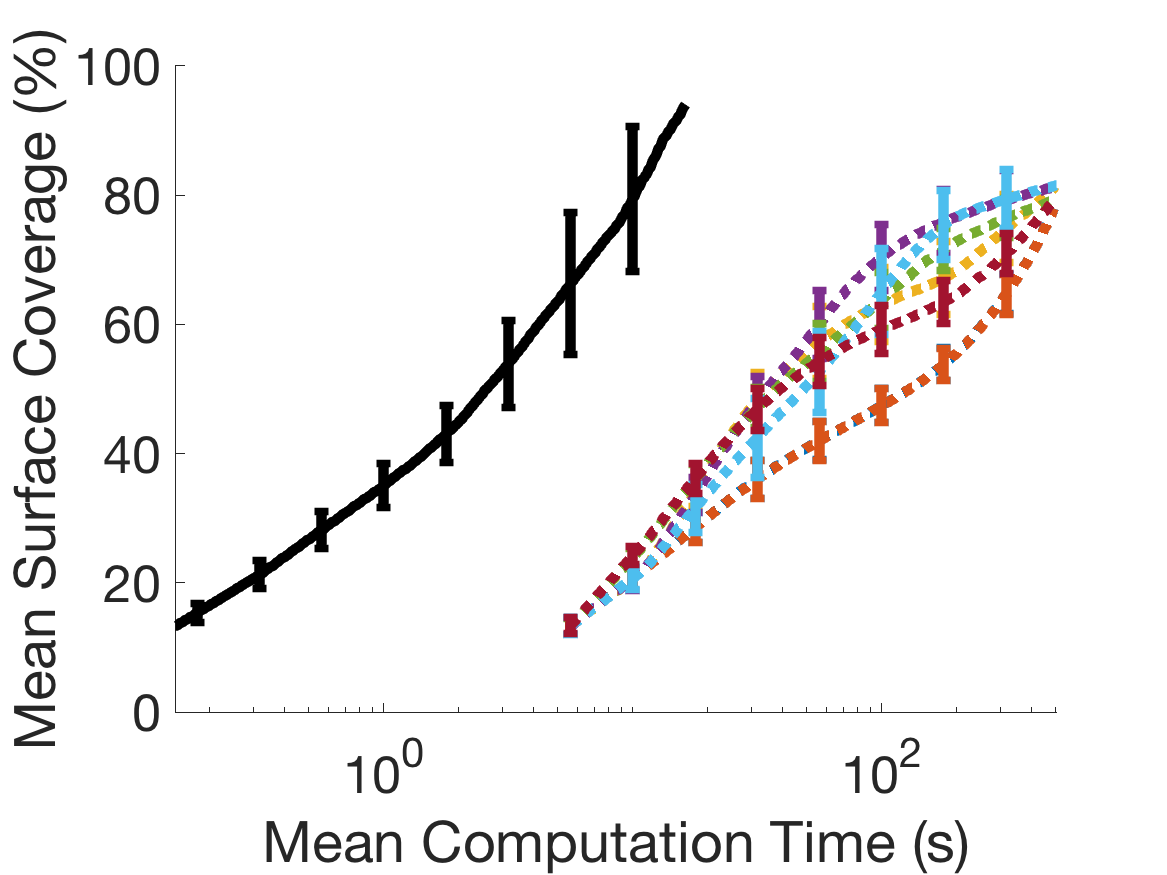}} \hfill
\subfloat[]{\includegraphics[width=.24\linewidth]{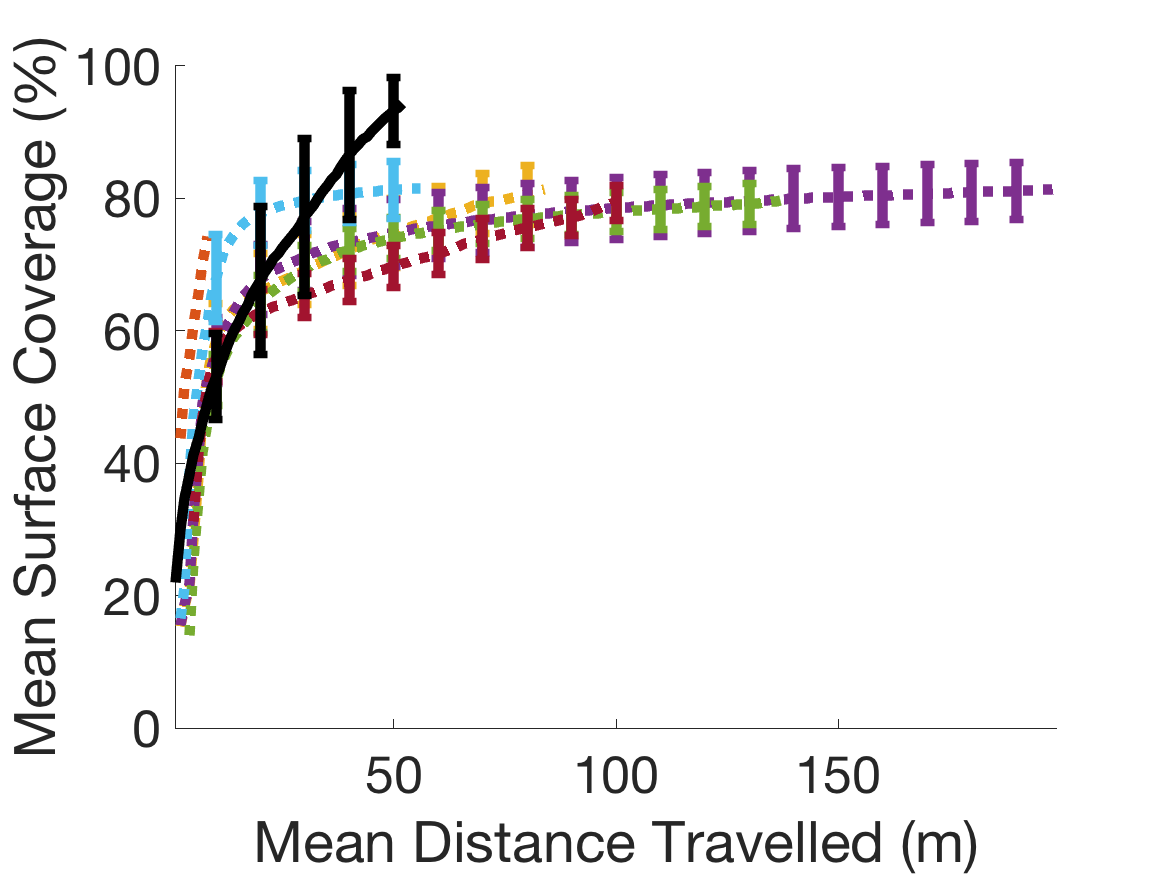}} \hfill
\captionsetup[subfigure]{}
\subfloat[Stanford Bunny ($1$~m) \citem{Turk1994}]{\includegraphics[width=.24\linewidth]{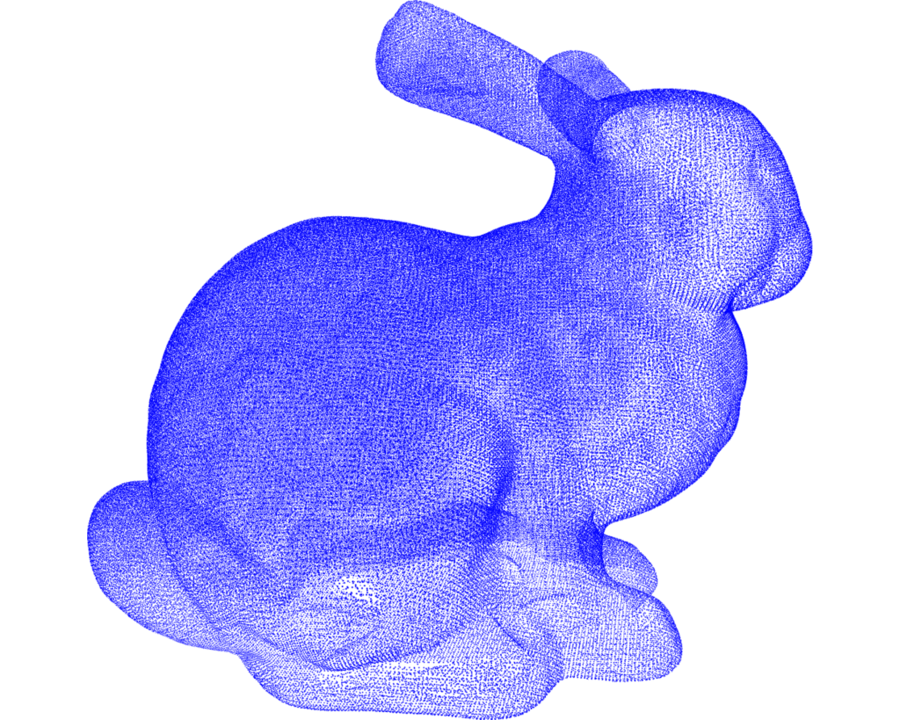}} \hfill
\captionsetup[subfigure]{labelformat=empty}
\subfloat[]{\includegraphics[width=.24\linewidth]{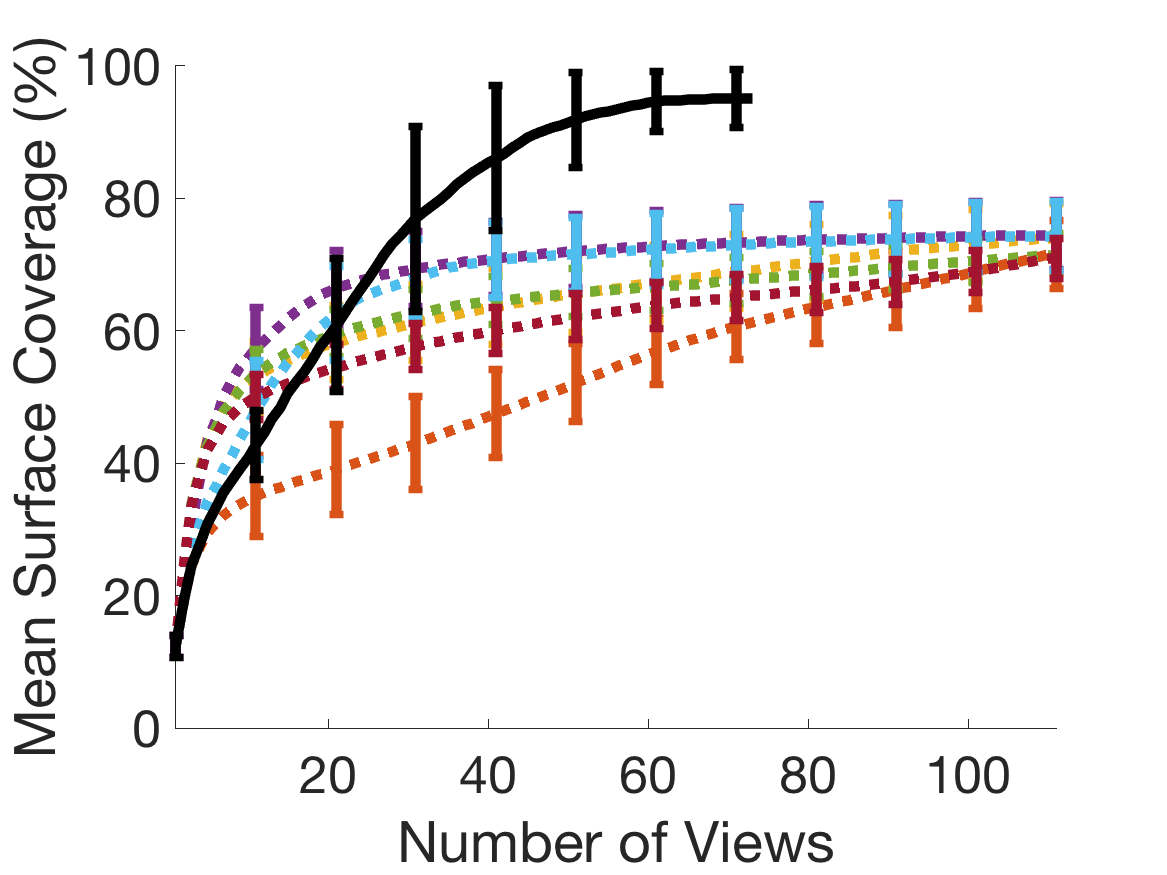}} \hfill
\subfloat[]{\includegraphics[width=.24\linewidth]{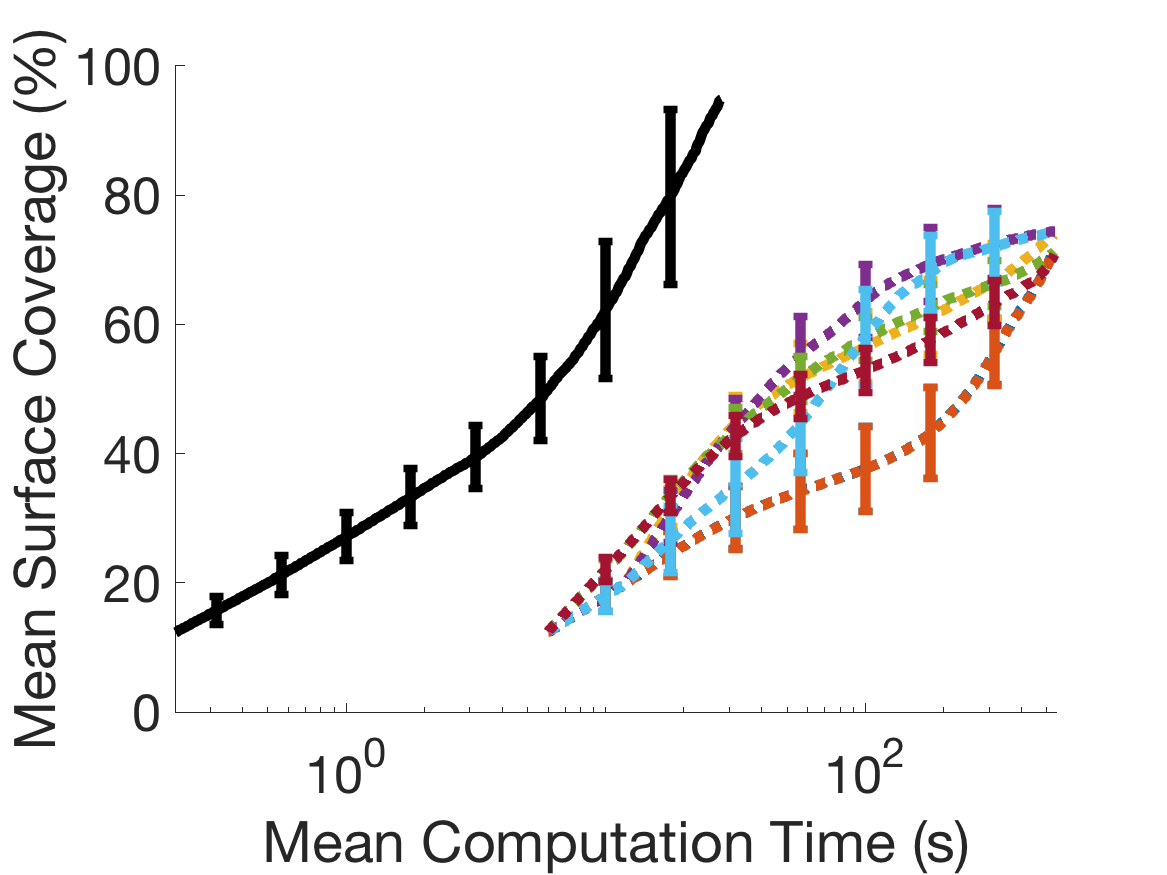}} \hfill
\subfloat[]{\includegraphics[width=.24\linewidth]{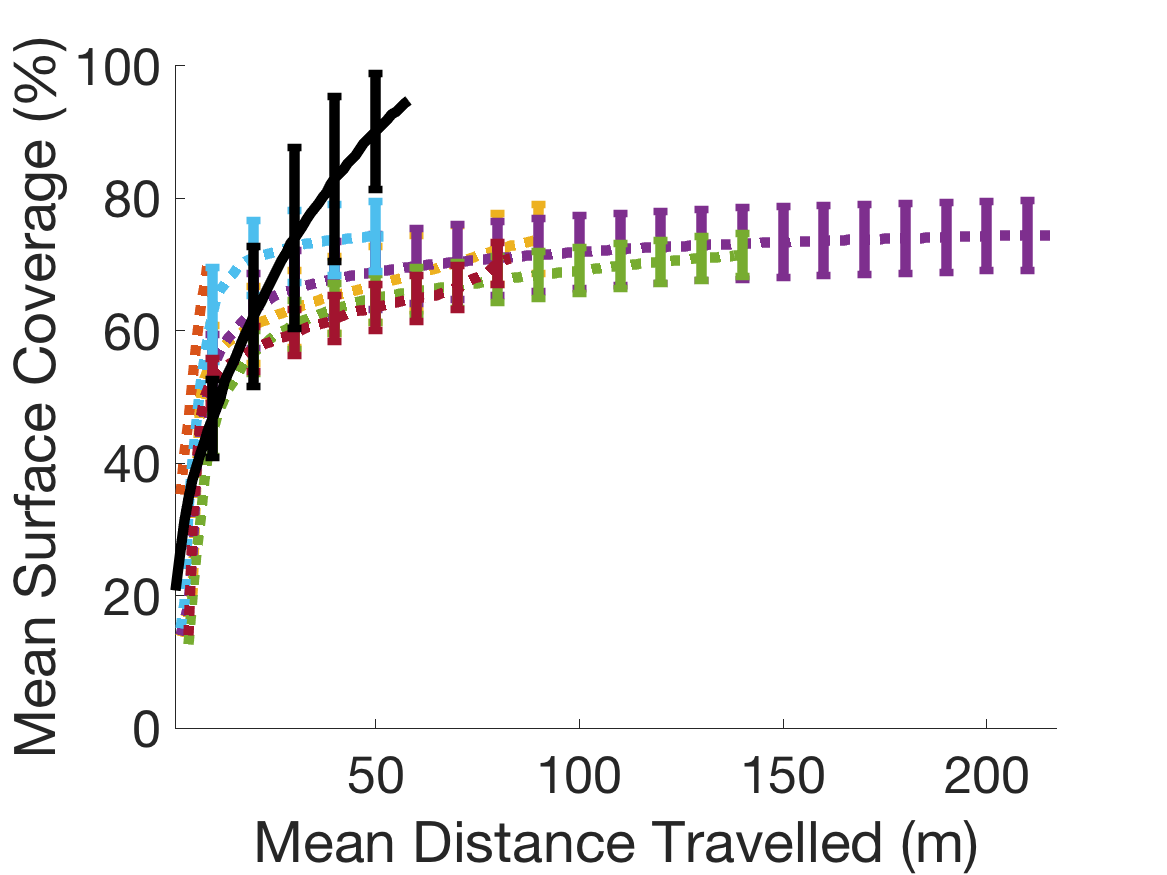}} \hfill
\captionsetup[subfigure]{}
\subfloat[Stanford Dragon ($1$~m) \citem{Curless1996}]{\includegraphics[width=.24\linewidth]{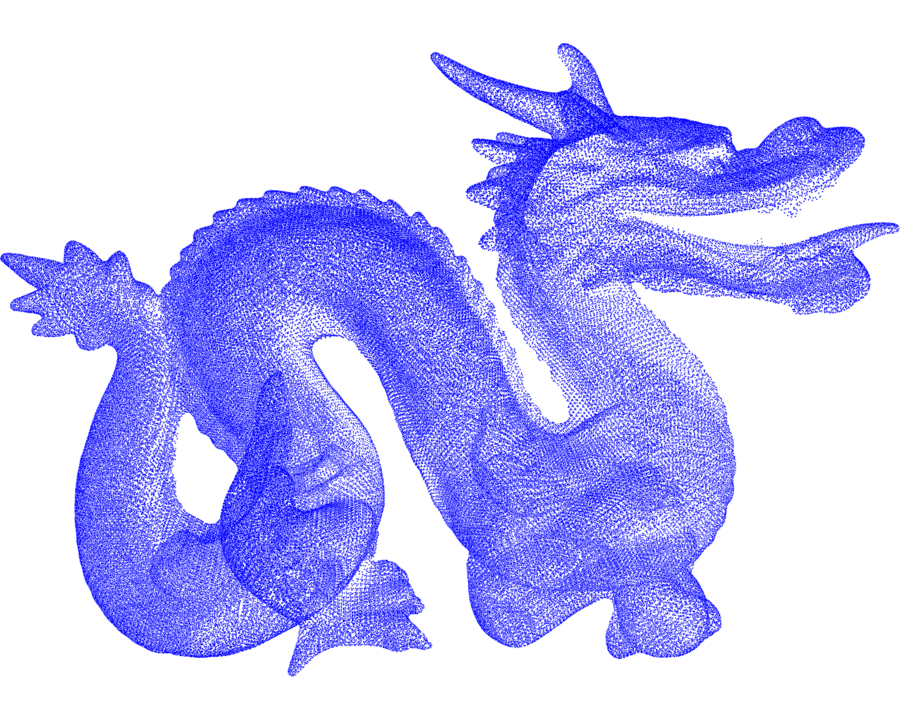}} \hfill
\captionsetup[subfigure]{labelformat=empty}
\subfloat[]{\includegraphics[width=.24\linewidth]{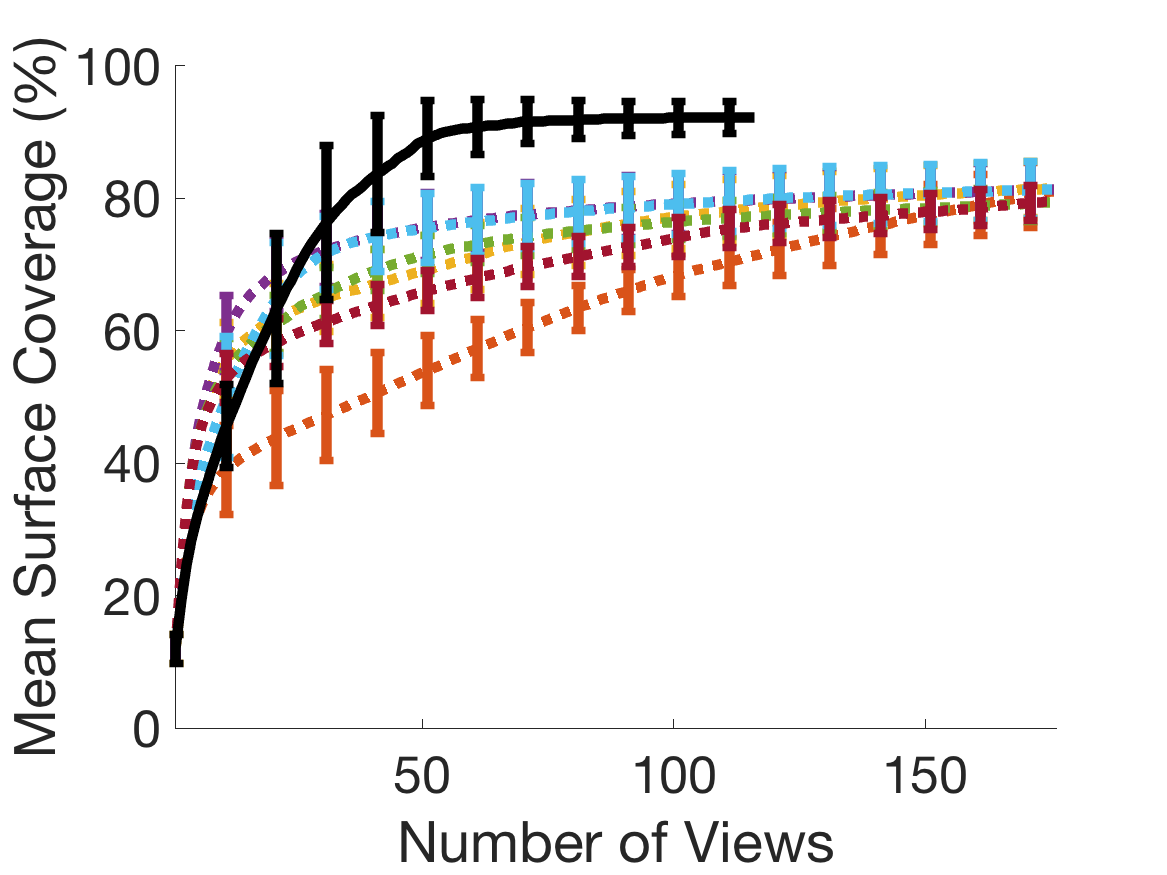}} \hfill
\subfloat[]{\includegraphics[width=.24\linewidth]{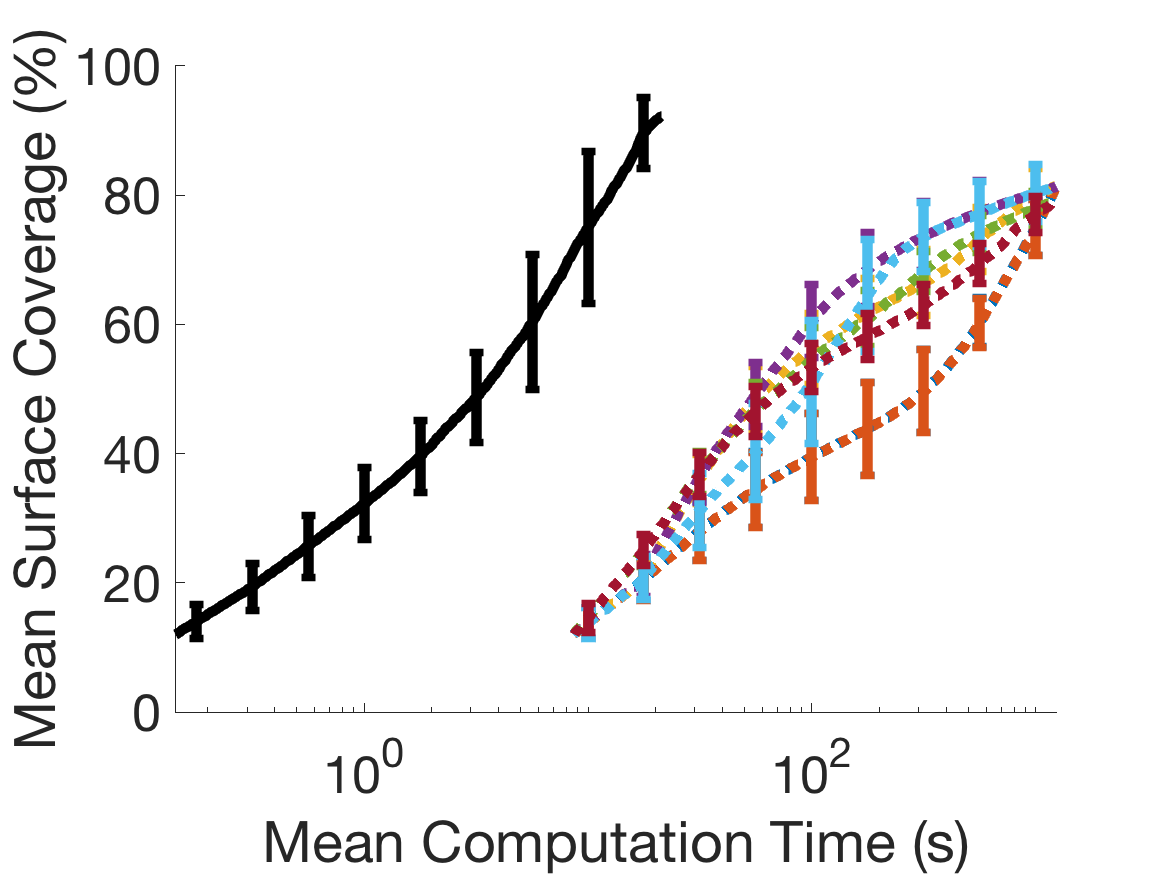}} \hfill
\subfloat[]{\includegraphics[width=.24\linewidth]{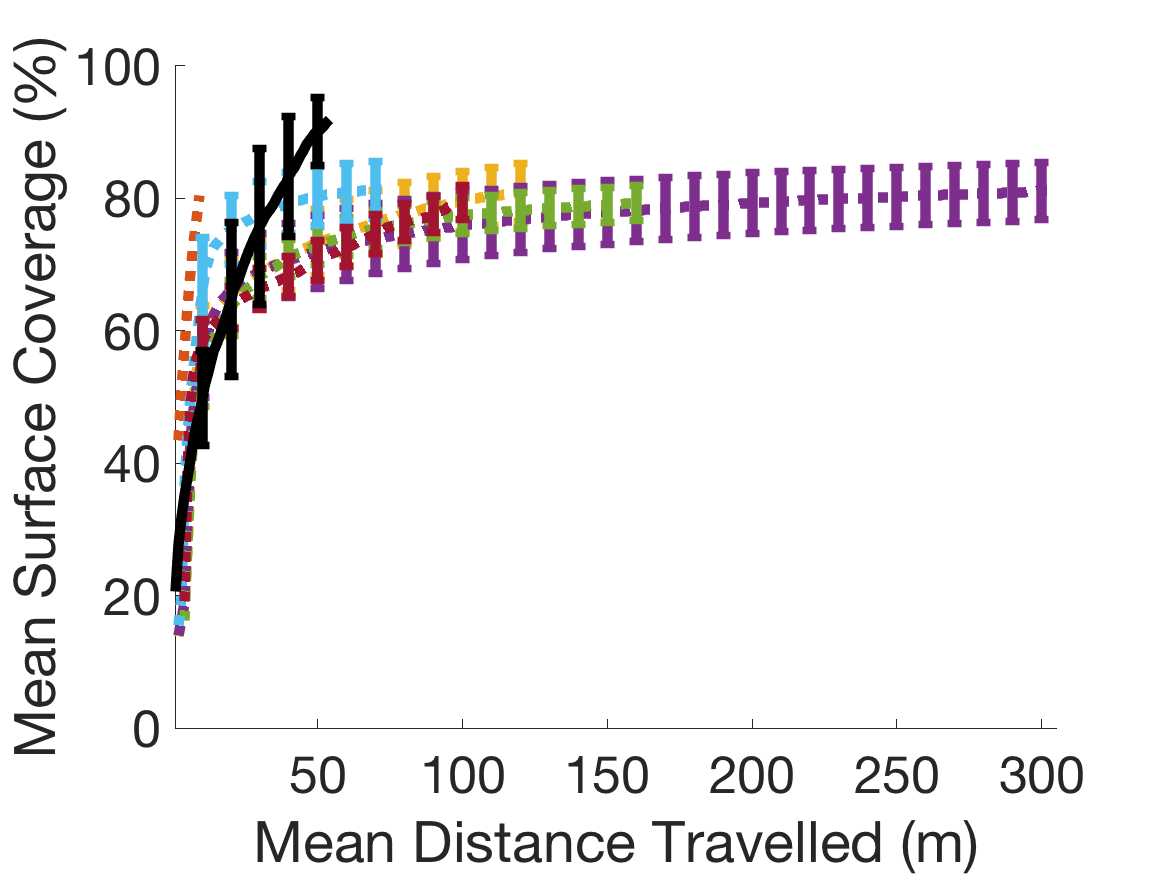}} \hfill
\captionsetup[subfigure]{}
\subfloat[Newell Teapot ($1$~m) \citem{Newell1975}]{\includegraphics[width=.24\linewidth]{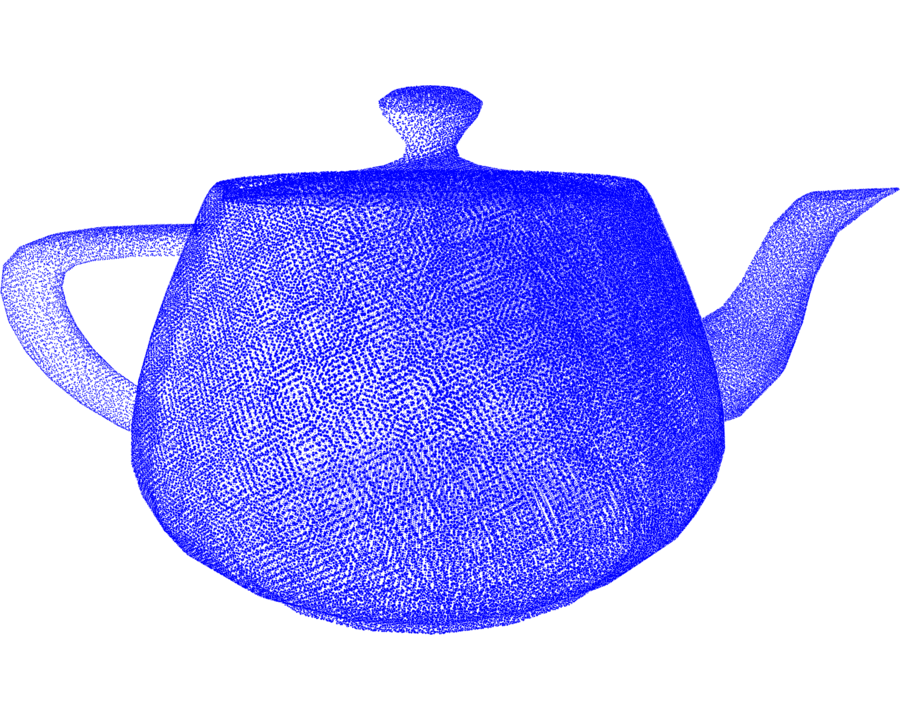}} \hfill
\captionsetup[subfigure]{labelformat=empty}
\subfloat[]{\includegraphics[width=.24\linewidth]{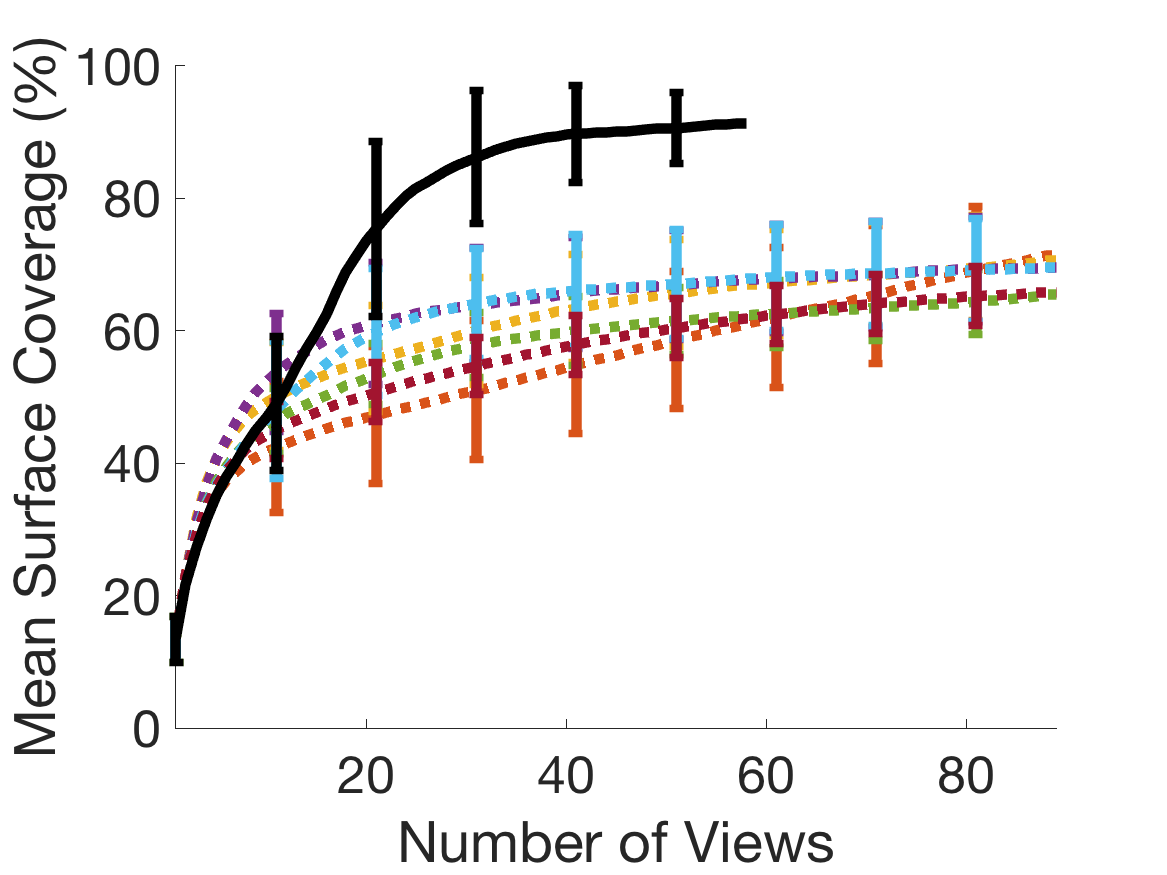}} \hfill
\subfloat[]{\includegraphics[width=.24\linewidth]{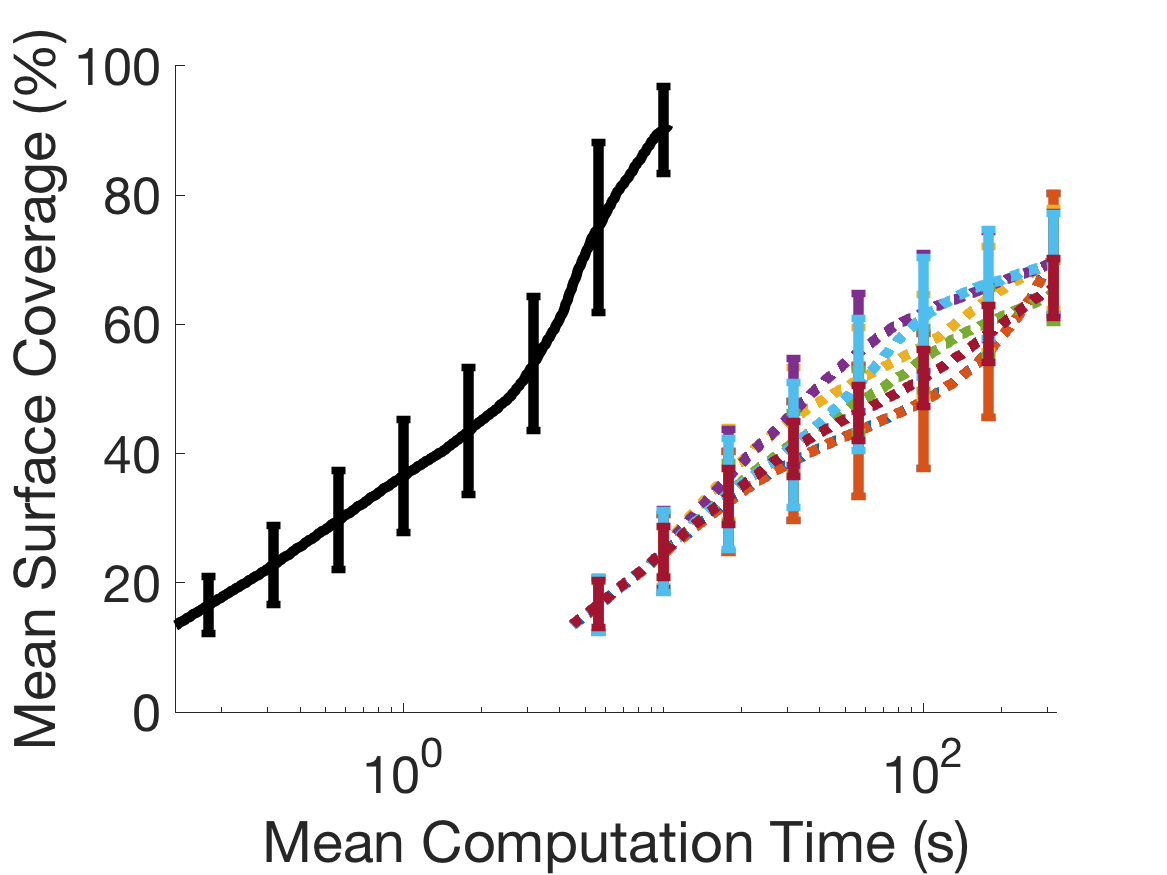}} \hfill
\subfloat[]{\includegraphics[width=.24\linewidth]{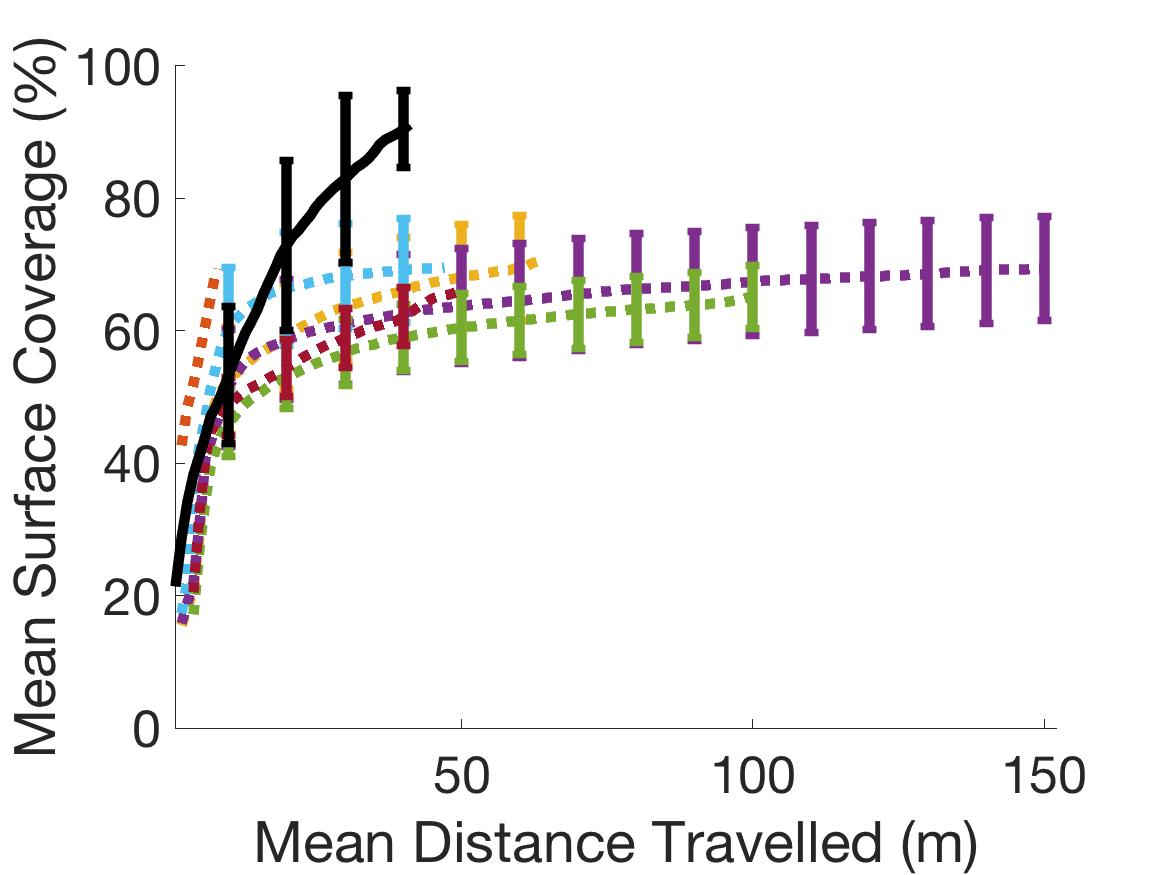}} \hfill
\captionsetup[subfigure]{}
\subfloat[Radcliffe Camera ($40$~m) \citem{Boronczyk2016}]{\includegraphics[width=.24\linewidth]{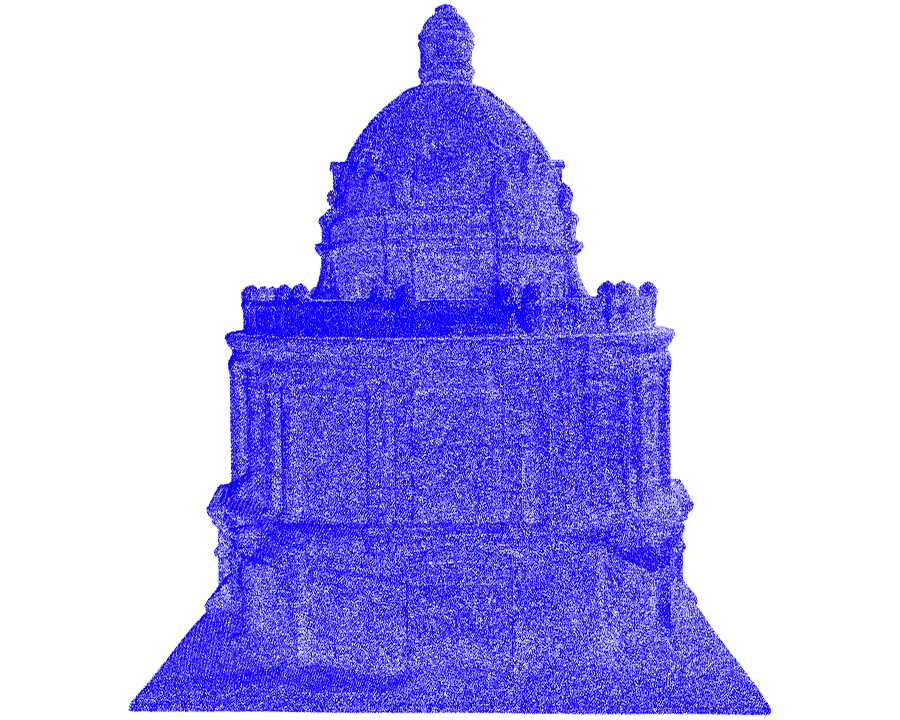}} \hfill
\captionsetup[subfigure]{labelformat=empty}
\subfloat[]{\includegraphics[width=.24\linewidth]{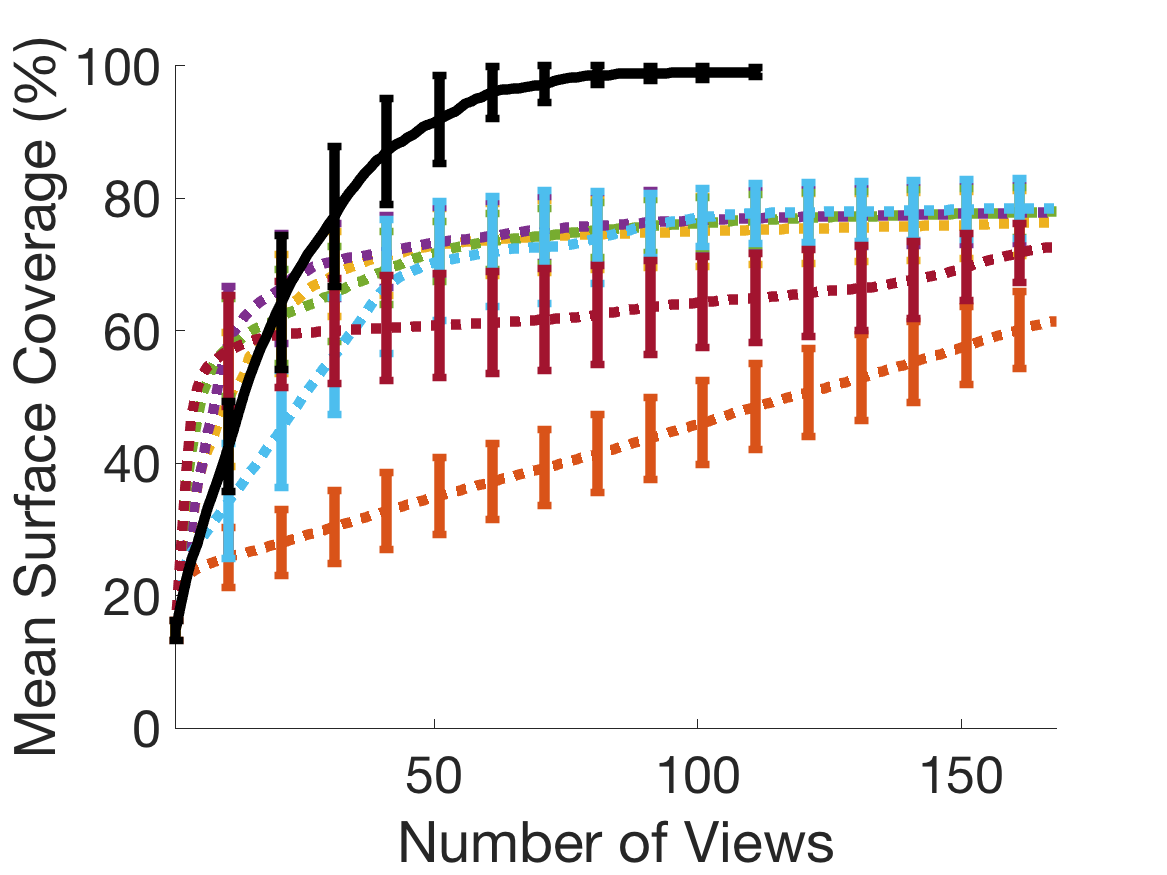}} \hfill
\subfloat[]{\includegraphics[width=.24\linewidth]{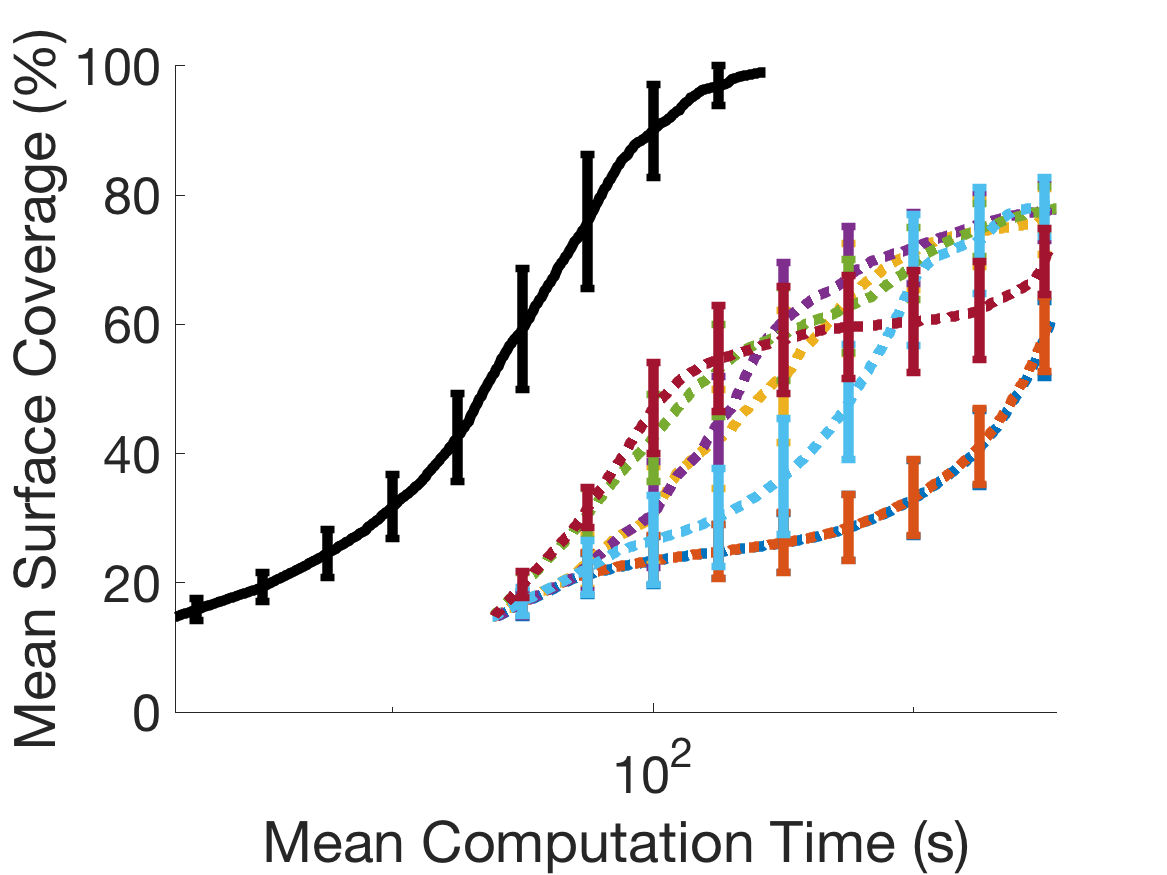}} \hfill
\subfloat[]{\includegraphics[width=.24\linewidth]{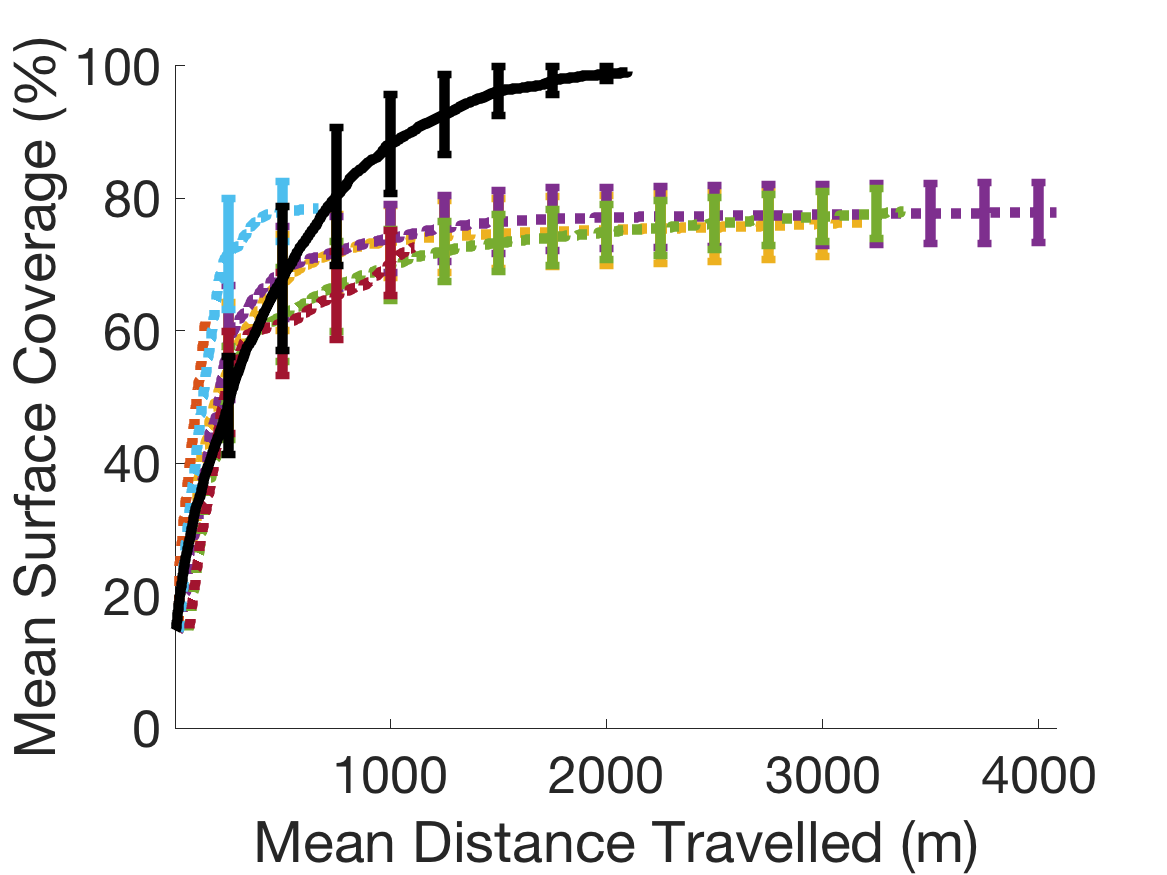}} \hfill
\subfloat[]{\includegraphics[width=.24\linewidth]{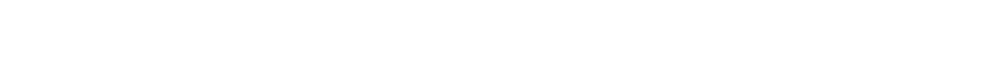}}
\subfloat[]{\includegraphics[width=.6\linewidth]{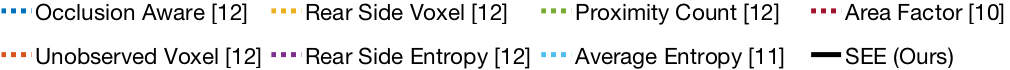}}
\caption{The performance of \gls{see} and state-of-the-art volumetric approaches \citem{Vasquez-Gomez2015, Kriegel2015, Delmerico2017} on four ($1$~m) standard models, the Stanford Armadillo \citem{Krishnamurthy1996a}, the Stanford Bunny \citem{Turk1994}, the Stanford Dragon \citem{Curless1996}, the Newell Teapot \citem{Newell1975} and on a full-scale ($40$~m) model of the Radcliffe Camera \citem{Boronczyk2016}. Noise-free measurements obtained by SEE are presented in the left-most column to illustrate the model. The graphs present the mean performance calculated from fifty independent trials on each model. Left to right they present the mean surface coverage vs the number of views, the mean computational time required to plan \gls{nbv}s and the mean distance travelled by the sensor. The error bars denote one standard deviation around the mean. These results show that SEE achieves higher surface coverage in less computational time and with near equivalent travel distances when compared to the evaluated volumetric approaches.}
\figlabel{results}
\end{figure*}

\gls{see} is compared to state-of-the-art \gls{nbv} approaches with volumetric representations, Area Factor (AF) \citem{Vasquez-Gomez2015}, Average Entropy (AE) \citem{Kriegel2015}, Occlusion Aware (OA) \citem{Delmerico2017}, Unobserved Voxel (UV) \citem{Delmerico2017}, Rear Side Voxel (RSV) \citem{Delmerico2017}, Rear Side Entropy (RSE) \citem{Delmerico2017} and Proximity Count (PC) \citem{Delmerico2017} on four standard models, the Stanford Armadillo \citem{Krishnamurthy1996a}, the Stanford Bunny \citem{Turk1994}, the Stanford Dragon \citem{Curless1996}, the Newell Teapot \citem{Newell1975} and on a full-scale model of the Radcliffe Camera \citem{Boronczyk2016}. The implementations of the volumetric approaches are provided by \cite{Delmerico2017}.

\subsection{Simulation Environment}

Measurements are simulated from a depth sensor by raycasting into a triangulated mesh of a scene model and adding Gaussian noise ($\mu = 0$~m, $\sigma = 0.01$~m) to the ray intersections to simulate a noisy 3D range sensor. These measurements are given to the \gls{nbv} algorithms as sensor observations. The process is repeated for each view requested by the algorithm.

The depth sensor is defined by a field-of-view in radians, $\alpha$, and a dimension in pixels, $w_\mathrm{x}$ and $w_\mathrm{y}$. The simulation environment contains no ground plane and the sensor can move unconstrained in three dimensions with six degrees of freedom. The sensor is prevented from moving inside scene surfaces by checking for intersections between the sensor path and the scene model. The sensor parameters used for the evaluation are $\alpha = \frac{\pi}{3}$~rad, $w_\mathrm{x} = 600$~px and $w_\mathrm{y} = 600$~px.

\subsection{Evaluation Parameters}

Potential views for the volumetric approaches are sampled from a given view surface (i.e., a view sphere) surrounding the scene as in \citem{Vasquez-Gomez2015, Delmerico2017}. Kriegel et al. \citem{Kriegel2015} does not restrict views to a view surface but we use the implementation provided by \citem{Delmerico2017} which does. The radius of the view sphere is defined as half the diagonal of the scene bounding box plus a chosen offset of $2$~m for the standard models and $16$~m for the Radcliffe Camera. The view distance for \gls{see} is set to the radius of the view sphere.

SEE uses a measurement density of $\rho = 4000$~points per m$^3$ for the standard models and $\rho = 60$~points per m$^3$ for the Radcliffe Camera. The resolution used is $r = 0.02$~m for the standard models and $r = 0.2$~m for the Radcliffe Camera. The volumetric approaches use the same resolutions for their voxel grids.

Every algorithm was run fifty times on each model for a given number of views. \gls{see} was run until its completion criterion was satisfied. The view limit for the \gls{infg} approaches on each model is set to $1.5\times$ the maximum number of views used by SEE to demonstrate their convergence. The number of views sampled on the view sphere is defined as $2.4\times$ the view limit as in \citem{Delmerico2017}. 

\subsection{Evaluation Metrics}

The algorithms are evaluated by calculating their relative surface coverage, computational time and sensor travel distance. These values are averaged across fifty experiments on each model \figref{results}.

\subsubsection{Surface Coverage}

The surface coverage of an approach is measured as the ratio of observed model points, $M_\mathrm{o}$, to total model points, $M_\mathrm{t}$,
  \[\tau := \frac{M_\mathrm{o}}{M_\mathrm{t}}\,.\]
A point is considered observed, $M_\mathrm{o} \subseteq M_\mathrm{t}$, if there is a measurement within $r_\mathrm{d}$ of the point. This registration distance is chosen as $r_\mathrm{d} = 0.005$~m for the standard models, as in \citem{Delmerico2017}, and $r_\mathrm{d} = 0.05$~m for the Radcliffe Camera model.

\subsubsection{Time}

The time taken to compute next best views is measured and added to a cumulative total. The time required for sensor travel is not considered.

\subsubsection{Distance}

The distance travelled by the sensor is measured by summing Euclidean distance between the positions of subsequent views.

\section{Discussion}

The experimental results demonstrate that \gls{see} outperforms the evaluated state-of-the-art volumetric approaches \figref{results} by requiring less computational time to plan views that obtain greater surface coverage with near equivalent travel distances, regardless of scene complexity and scale. \gls{see} is shown to consistently obtain high surface coverage for models with different surface complexities and scales while the volumetric approaches demonstrate varying performance. 

Standard models with a large amount of self-occlusions (e.g., the ears of the Stanford Bunny and the handle of the Newell Teapot) demonstrate the advantages of the adaptable views used by \gls{see}. The evaluated volumetric approaches perform worse on these problems as they do not adjust their views to account for occlusions. The view selection metric presented in \citem{Kriegel2015} does adapt views to handle occlusions but this is not included in the implementation provided by \citem{Delmerico2017}. 

The Radcliffe Camera model demonstrates the difficulty of scaling volumetric approaches to large scenes. The large resolution necessary for reasonable raytracing allows voxels to be observed by discontinuous measurements \figref{marquee}.

The experiments show that the computational performance of \gls{see} is logarithmically better than the volumetric approaches. The poor performance of the volumetric approaches is due to the computational complexity of raytracing a high-resolution voxel grid from every view on the view sphere when selecting a \gls{nbv}. The limited scalability of the volumetric approaches with scene size is demonstrated by the difference in computational performance between the standard models and the Radcliffe Camera model.   

While \gls{see} travels a larger distance per-view in the experiments, it initially achieves equivalent surface coverage per unit distance. The volumetric approaches then appear to continue to travel without significantly improving coverage while \gls{see} continues to increase coverage as it travels. As a result, by the time \gls{see} terminates it has travelled distances equivalent to many of the other approaches but has achieved higher surface coverage.

\section{Conclusion}

\gls{see} is a scene-model-free approach to NBV planning that uses a density representation. The representation defines a \textit{frontier} between fully and partially observed surfaces based on a user-specified resolution and measurement density. View proposals are generated to observe this frontier and extend the scene coverage. \gls{nbv}s are selected and new measurements are obtained until the scene is fully observed with the given measurement density and at the specified resolution. 

The density representation used by \gls{see} has a number of advantages over volumetric and surface representations. Unlike volumetric representations, the complexity of \gls{see} only scales with the number of measurements and not scene scale, making it possible to obtain high-resolution models of large scenes. In contrast to many surface approaches the measurement density and resolution parameters can be specified intuitively and only a single survey stage is required.  

Experimental results show that \gls{see} outperforms state-of-the-art volumetric approaches in terms of surface coverage and computation time. It take less computation time to propose views that achieve greater surface coverage with an equivalent travel distance. 

SEE was only compared to publicly available volumetric approaches as we were unable to attain implementations of relevant surface approaches. We plan to implement state-of-the-art surface (e.g., \cite{Dierenbach2016}) and/or combined approaches (e.g., \cite{Kriegel2015}) and present comparisons with these in future work. SEE may be made available to other researchers upon request to facilitate comparisons. We are also working to deploy and test \gls{see} on real-world problems with an aerial platform.

\renewcommand*{\bibfont}{\footnotesize}
{\renewcommand{\markboth}[2]{}
 \printbibliography}

\end{document}